\documentclass{article}

\usepackage{microtype}
\usepackage{graphicx}
\usepackage{subfigure}
\usepackage{booktabs} 

\usepackage{hyperref}

\usepackage[accepted]{icml2024}

\usepackage{amsmath,amsfonts,bm,amssymb,mathtools,amsthm}
\usepackage{color,xcolor,xspace}
\usepackage{booktabs}
\usepackage{thm-restate}

\newcommand{\bb}[1]{{\mathbb{#1}}}

\def\eqref#1{Eq.~(\ref{#1})}

\def\1{\bm{1}}

\newcommand{\defeq}{\vcentcolon=}

\def\rveps{{\boldsymbol{\epsilon}}}

\def\rvmu{{\boldsymbol{\mu}}}

\def\rvv{{\mathbf{v}}}

\def\rvx{{\mathbf{x}}}

\def\rmI{{\mathbf{I}}}

\DeclareMathAlphabet{\mathsfit}{\encodingdefault}{\sfdefault}{m}{sl}
\SetMathAlphabet{\mathsfit}{bold}{\encodingdefault}{\sfdefault}{bx}{n}

\def\gL{{\mathcal{L}}}

\def\gN{{\mathcal{N}}}

\newcommand{\ptheta}{p_\theta}
\newcommand{\pphi}{p_\phi}
\newcommand{\qlambda}{q_\lambda}
\newcommand{\ttm}{\rvx_t|\rvx_{t-1}}

\newcommand{\tmtz}{\rvx_{t-1}|\rvx_{t},\rvx_1}
\newcommand{\tmz}{\rvx_{t-1}|\rvx_{1}}
\newcommand{\tz}{\rvx_{t}|\rvx_{1}}

\newcommand{\balpha}{\bar\alpha}
\usepackage{amsmath}
\usepackage{amssymb}
\usepackage{mathtools}
\usepackage{amsthm}

\usepackage[capitalize,noabbrev]{cleveref}

\usepackage[textsize=tiny]{todonotes}
\definecolor{mygreen}{rgb}{0.0, 0.5, 0.0}
\definecolor{myyellow}{rgb}{0.85, 0.57, 0.0}
\definecolor{myred}{rgb}{0.8, 0.0, 0.0}

\newcommand{\grayline}[1]{\multicolumn{3}{c|}{{\color{gray} -------} #1 {\color{gray}-------}}}
\newcommand{\graylineempty}[1]{\multicolumn{3}{c}{{\color{gray} -------} #1 {\color{gray}-------}}}

\usepackage{tabularx}
\newcolumntype{C}{>{\centering\arraybackslash}X}
\usepackage{ulem}
\usepackage{scrextend}
\usepackage{wrapfig}
\usepackage{algorithm}
\usepackage{algpseudocode}
\usepackage{multirow} 

\usepackage{enumitem}

\newcommand{\modelnamenew}{EDDPM\xspace}
\newcommand{\modelnamenews}{EDDPMs\xspace}

\icmltitlerunning{Generalized Encoding-Decoding Diffusion Probabilistic Models}

\begin{document}

\twocolumn[
\icmltitle{Unified Generation, Reconstruction, and Representation:\\
Generalized Diffusion with Adaptive Latent Encoding-Decoding
}
\vspace{-0.5cm}

\icmlsetsymbol{equal}{*}

\begin{icmlauthorlist}
\icmlauthor{Guangyi Liu}{equal,mbzuai}
\icmlauthor{Yu Wang}{equal,ucsd}
\icmlauthor{Zeyu Feng}{equal,ucsd}
\icmlauthor{Qiyu Wu}{tu}
\icmlauthor{Liping Tang}{mbzuai}
\icmlauthor{Yuan Gao}{su}

\icmlauthor{Zhen Li}{cuhk}
\icmlauthor{Shuguang Cui}{cuhk}
\icmlauthor{Julian McAuley}{ucsd}
\icmlauthor{Zichao Yang}{cmu}

\icmlauthor{Eric P. Xing}{mbzuai,cmu}
\icmlauthor{Zhiting Hu}{ucsd}
\end{icmlauthorlist}
\icmlaffiliation{mbzuai}{MBZUAI}
\icmlaffiliation{ucsd}{UC San Diego}
\icmlaffiliation{tu}{University of Tokyo}
\icmlaffiliation{su}{Stanford University}
\icmlaffiliation{cuhk}{CUHK-Shenzhen}
\icmlaffiliation{cmu}{CMU}

\icmlcorrespondingauthor{Guangyi Liu}{guangyiliu.xx@gmail.com}
\icmlkeywords{Machine Learning, ICML}

\vskip 0.2in
]



\printAffiliationsAndNotice{\icmlEqualContribution} 


\begin{abstract}

The vast applications of deep generative models are anchored in three core capabilities---{\it generating} new instances, {\it reconstructing} inputs, and learning compact {\it representations}---across various data types, such as discrete text/protein sequences and continuous images. 
Existing model families, like variational autoencoders (VAEs), generative adversarial networks (GANs), autoregressive models, and (latent) diffusion models, generally excel in specific capabilities and data types but fall short in others. 
We introduce {\it Generalized  \uline{E}ncoding-\uline{D}ecoding \uline{D}iffusion \uline{P}robabilistic \uline{M}odels} (\modelnamenews) which integrate the core capabilities for broad applicability and enhanced performance. 
\modelnamenews generalize the Gaussian noising-denoising in standard diffusion by introducing parameterized encoding-decoding. Crucially, \modelnamenews are compatible with the well-established diffusion model objective and training recipes, allowing effective learning of the encoder-decoder parameters {\it jointly} with diffusion. By choosing appropriate encoder/decoder (e.g., large language models), \modelnamenews naturally apply to different data types.
Extensive experiments on text, proteins, and images demonstrate the flexibility to handle diverse data and tasks and the strong improvement over various existing models. Code is available at {\url{https://github.com/guangyliu/EDDPM}}

\end{abstract}


\begin{figure}[!t]
    \centering
    \includegraphics[width=\columnwidth]{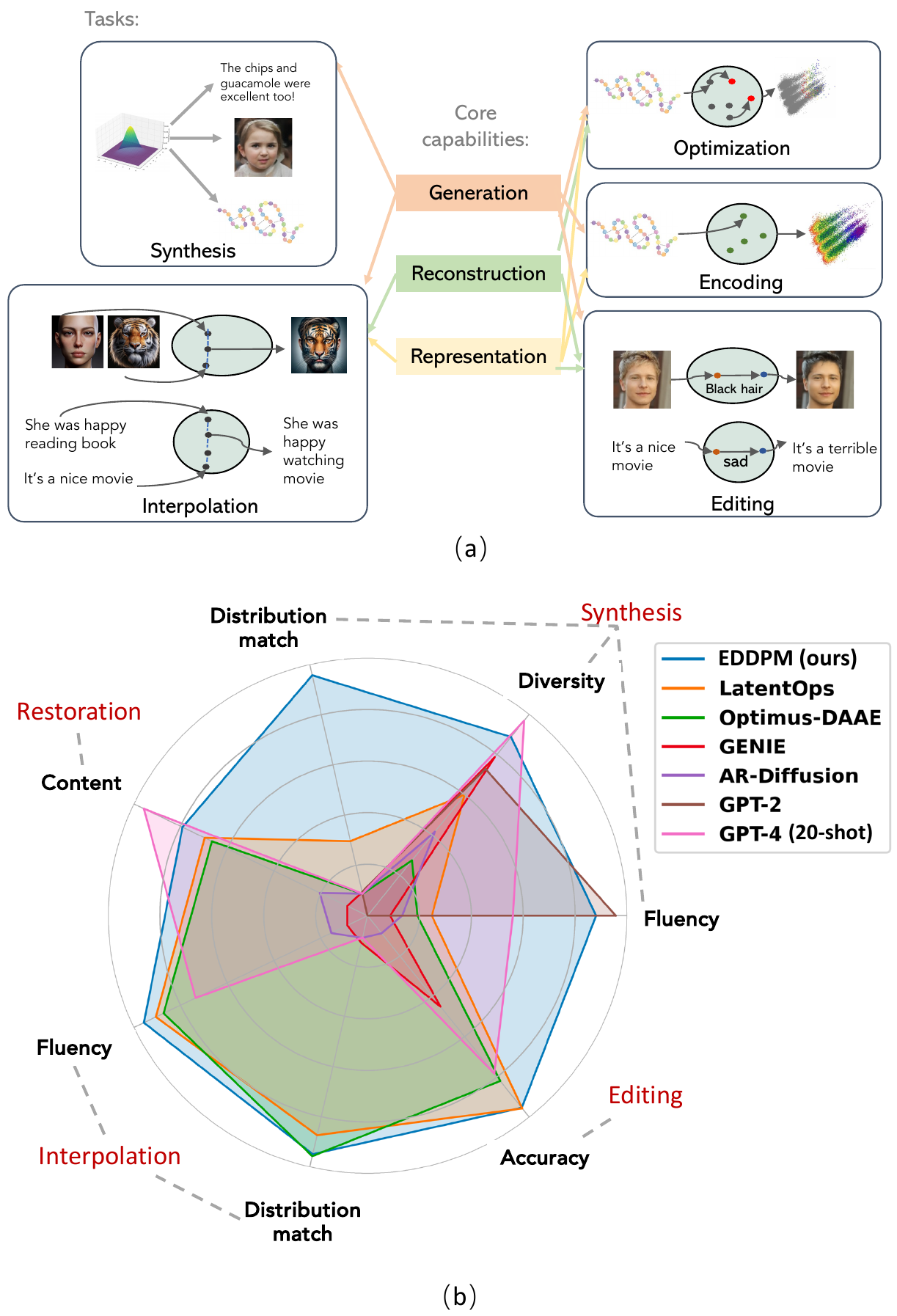}
    \vspace{-0.6cm}
    \caption{{\bf (a)} Generation, reconstruction, and representation are the core capabilities for diverse applications. {\bf (b)} \modelnamenew shows comprehensive abilities on different text tasks in the customer-review domain (\S\ref{sec:exp_text}).
    }
    \label{fig:intro}
    \label{fig:radar_text}
    \vspace{-0.4cm}
\end{figure}

\vspace{-0.4cm}
\section{Introduction}
\label{sec:introduction}

Numerous real-world applications involve synthesizing, modifying, restoring, and encoding data of different types, such as text, images, and biological molecules. Deep generative models are pivotal in these applications, owing to their three fundamental capabilities: (1) {\it Generation} of new samples from the data distribution; (2) {\it Reconstruction} of given instances with high fidelity; and (3) Extracting compact {\it representations} of raw data. Different applications can require different combinations of the capabilities (Figure~\ref{fig:intro}a): for example, protein editing requires to produce a new valid protein sequence ({\it generation}) that retains many properties of the original sequence ({\it reconstruction}), while text interpolation could benefit from a latent space for continuous transitioning ({\it representation}) before mapping back to the discrete text space ({\it generation}). 

\begin{figure*}[t]
    \centering
    \includegraphics[width=\textwidth]{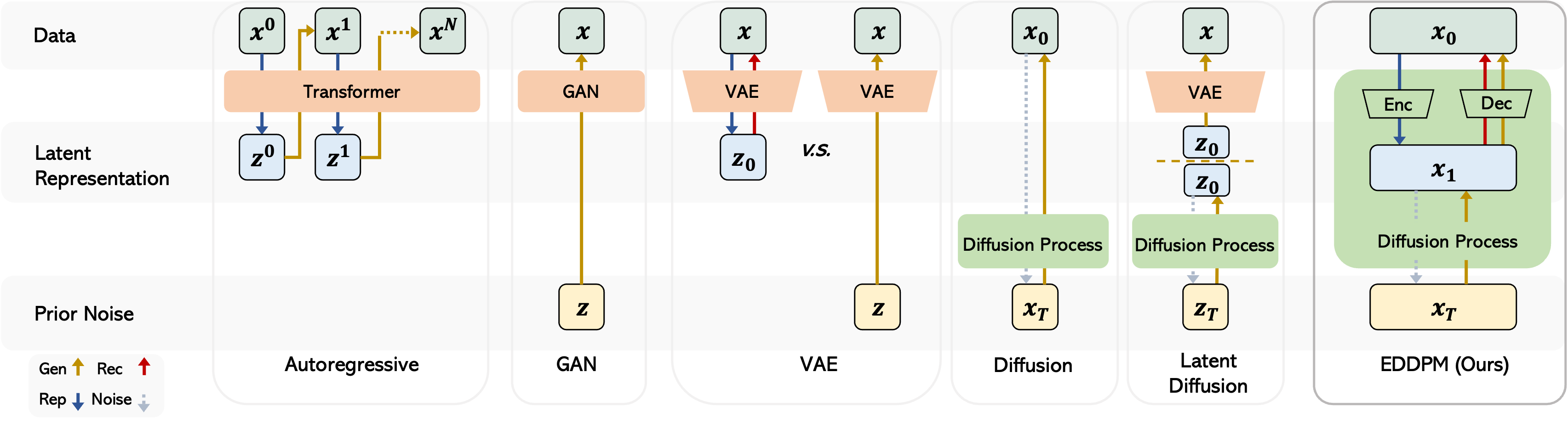}
    \vspace{-0.8cm}
    \caption{
    Different families of deep generative models.
    }
    \label{fig:model}
\end{figure*}

Existing deep generative models  typically show strengths in some, but not all, of the three capabilities, resulting in limited applicability or suboptimal performance (Figure~\ref{fig:model}). For example, variational autoencoders \citep[VAEs,][]{kingma2013auto} are known for their tradeoff between realistic generation and faithful reconstruction \citep{DBLP:conf/iclr/0022KSDDSSA17,Higgins2016betaVAELB}, especially on text sequences \citep{DBLP:conf/conll/BowmanVVDJB16,yang2017improved, DBLP:conf/emnlp/LiGLPLZG20,liu2022dont,liu2022composable}; Generative adversarial networks \citep[GANs,][]{goodfellow2014generative} inherently lack inference of latent representations. Despite the rich subsequent research of incorporating latent inference \citep{DBLP:conf/ijcnn/ZhouGH19, DBLP:journals/pami/XiaZYXZY23,DBLP:journals/corr/DonahueKD16,ALI,ALICE_Advasarial} 
such as GAN inversion \citep{GAN-inversion-survey, stylegan-inversion}, they fall short of faithfully reconstructing the inputs \citep{diffae}; Autoregressive models (e.g., modern large language models) excel in generating text, but often with limited diversity and lacking compact semantic representations \citep{Brown2020LanguageMA}. Similarly, recent diffusion models \citep{DBLP:conf/icml/Sohl-DicksteinW15,DBLP:conf/nips/HoJA20, song2021scorebased} deliver new state-of-the-arts on synthesizing photo-realistic images, yet lack compact data representation and often rely on separately trained components for remediation \citep{Rombach2021HighResolutionIS,diffae}.

Moreover, these models often struggle with different forms of data. For example, it has been notoriously difficult for diffusion models and GANs to deal with text and protein data due to their discrete nature \citep{li2022diffusionlm, qin2022cold, kusner2016gans, yu2017seqgan, de2021survey}, and VAEs suffer from ``posterior collapse'' on text sequences \citep{DBLP:conf/conll/BowmanVVDJB16,yang2017improved}. 

In this paper, we investigate a new deep generative approach based on {\it generalized} diffusion, that inherently integrates the three core capabilities and offers added flexibility to model both discrete and continuous data. 
Specifically, starting with the popular variational inference perspective of diffusion models \citep[as in DDPM,][]{DBLP:conf/nips/HoJA20}, we show that the standard formulation can be generalized by plugging in arbitrary ``noising'' operation at early diffusion steps, as long as the inverse ``denoising'' operation can be modeled. We can thus go beyond the common Gaussian noise \citep[or other predefined image degradation,][]{DBLP:journals/corr/abs-2208-09392} and introduce {\it parameterized encoder-decoder} in place of noising-denoising. 
Crucially, the encoder-decoder parameters can naturally be learned together with other diffusion parameters using the original DDPM framework which is well-established for stable and scalable training \citep{Rombach2021HighResolutionIS}. 

More specifically, we introduce encoder-decoder at the first diffusion step. That is, the diffusion process first encodes the input data into a low-dimensional latent vector, followed by common Gaussian noising steps. 
The resulting approach, 
\modelnamenews (Generalized \uline{E}ncoding-\uline{D}ecoding \uline{D}ffusion \uline{P}robabilistic \uline{M}odels), 
combine a number of desirable properties and overcome difficulties in aforementioned deep generative models:
\begin{enumerate}[label={{\bf (\arabic*)}},itemsep=3pt,parsep=0pt,topsep=0pt]
    \item Like VAEs and latent diffusion models (which are based on VAEs), \modelnamenews offer compact semantic {\it representation} of data. Moreover, thanks to the unified learning of the encoder-decoder within the standard diffusion framework, \modelnamenews obtain a much {\it improved} representation space than VAEs and latent diffusion, as shown in \S\ref{sec:experiments}.

    \item The improved representation space also allows \modelnamenews to avoid the trade-off between {\it generation} and {\it reconstruction} capabilities observed in VAEs and augmented GANs \citep{alice}. Therefore, \modelnamenews seamlessly integrate the three core capabilities, leading to more diverse applications and enhanced performance.

    \item The flexibility of the generalized diffusion formulation allows us to specify any desired encoder-decoder for modeling both {\it discrete} and {\it continuous} data. \modelnamenews thus overcome the difficulty of standard diffusion and GANs on text and other discrete modalities.

    \item Moreover, as described in \S\ref{sec:experiments}, we could further plug in large {\it pretrained} (autoregressive) language models (LMs) for initializing the encoder-decoder. This leads to greatly improved performance than previous text diffusion models \citep{li2022diffusionlm, genie, ardiffusion} that are inherently incompatible with off-the-shelf pretrained LMs.  
\end{enumerate}

We conduct extensive experiments on text, images, and protein sequences. \modelnamenews demonstrate comprehensive capabilities across a wide range of tasks, such as data synthesis, reconstruction, interpolation, editing, and optimization (e.g., Figure~\ref{fig:intro}b) on the different data modalities. 

The unified capabilities and enhanced performance of \modelnamenews compared to existing generative models demonstrate its significant potential as a foundational technique for developing new, broadly applicable foundation models.

\section{Background}
\label{sec:background}
We first give brief background of the popular diffusion model formulation from the variational inference perspective \citep{DBLP:conf/icml/Sohl-DicksteinW15,DBLP:conf/nips/HoJA20, DDIM}. The training objective and model configurations have been well-established and are effective for stable and scalable training in practice~\citep{Rombach2021HighResolutionIS}, showing advantages to alternative diffusion formalisms and other generative models. Our approach (\S\ref{sec:method}) fits seamlessly in the (generalized) formulation and inherits these advantages.

A standard diffusion model consists of {\it noising} and {\it denoising} processes. Starting with a raw data point $\mathbf{x}_0$, the noising process sequentially adds Gaussian noise to the data at each diffusion step $t\in\{1,\dots,T\}$, following:
\begin{equation}
\small
\begin{split}
q(\mathbf{x}_t|\mathbf{x}_{t-1}) &:= \mathcal{N}(\mathbf{x}_t; \sqrt{1 - \beta_t} \mathbf{x}_{t-1}, \beta_t \mathbf{I}), 
\end{split}
\label{eq:ddpm_forward}
\end{equation}
where $\beta_t$ is the predefined variance schedule, and $\mathbf{x}_T$ becomes a standard Gaussian noise vector as $T\to \infty$. The denoising process learns to invert the above by learning a denoising operation $p_\theta(\rvx_{t-1}|\rvx_{t})$ at each step $t$. When $\beta_t$ is sufficiently small, we assume a Gaussian form:
\begin{equation}
\small
\begin{split}
p_\theta(\rvx_{t-1}|\rvx_{t}) &:=\gN(\rvx_{t-1};\boldsymbol{\mu}_\theta(\rvx_t,t),\mathbf\Sigma_\theta(\rvx_t,t)).
\end{split}
\label{eq:reverse}
\end{equation}
The parameters $\theta$ are learned with a variational lower bound on the marginal likelihood: with a Gaussian prior $p(\rvx_T)$,
\begin{equation}
\small
\begin{split}
    {\mathcal{L}}(\theta) &= \mathbb{E}_{q}[-\log p_\theta(\mathbf{x}_0)] \\
    &\leq \mathbb{E}_{q}\left[-\log p(\rvx_T) - \sum_{t=1}^{T}\log\frac{p_\theta(\rvx_{t-1}|\rvx_t)}{q(\rvx_t|\rvx_{t-1})}\right].
\end{split}
    \label{eq:elbo}
\end{equation}

Recent diffusion formulation, in particular the denoising diffusion probabilistic models (DDPMs)  \citep{DBLP:conf/icml/Sohl-DicksteinW15,DBLP:conf/nips/HoJA20}, introduces a series of derivations to simplify the above objective, by splitting the sum and transforming and cancelling out relevant terms. The lower bound is thus rewritten as:
\begin{equation}
\footnotesize
\begin{split}
   \mathbb{E}_{q}\bigg[& -\log p(\rvx_T) - \log \frac{p_\theta(\rvx_{0}|\rvx_1)}{q(\rvx_1|\rvx_{0})} \bigg.\\
   &- \sum_{t=2}^{T}\log\frac{p_\theta(\rvx_{t-1}|\rvx_t)}{q(\rvx_{t-1}|\rvx_{t},\rvx_0)} \frac{q(\rvx_{t-1}|\rvx_0)}{q(\rvx_t|\rvx_0)} \bigg] \\
   \vspace{5pt}\\
    =\  &\mathbb{E}_{q} [ \underbrace{\text{KL}\left( q(\rvx_T|\rvx_0) || p(\rvx_T) \right)}_{{\mathcal{L}}_T} -\underbrace{\log p_\theta(\rvx_0|\rvx_1)}_{{\mathcal{L}}_{0}} ] \\
    & \quad + \sum_{t = 2}^T \underbrace{\text{KL}\left( q(\rvx_{t-1}|\rvx_t, \rvx_0) || p_\theta(\rvx_{t-1}|\rvx_t) \right)}_{{\mathcal{L}}_{t-1}}.
\end{split}
    \label{eq:ddpm_loss_kl}
\end{equation}

{\citet{DBLP:conf/nips/HoJA20} further proposed to reparameterize $p_\theta(\rvx_{t-1}|\rvx_{t})$ in \eqref{eq:reverse} using $\boldsymbol{\mu}_\theta(\rvx_t,t) = \frac{1}{\sqrt{\bar{\alpha}_t}}(\mathbf{x}_t - \sqrt{1-\bar\alpha_t} \rveps_\theta(\rvx_t, t))$, }
where $\alpha_t$ and $\bar{\alpha}_t$ are weights defined by $\beta_t$.
{The model is trained to directly predict the noise term $\rveps_\theta(\rvx_t, t)$ instead of the mean $\boldsymbol{\mu}_\theta(\rvx_t,t)$, which has demonstrated effectiveness in practice.}

Despite their strengths, diffusion models have several limitations. The noisy nature of the latent variables $\rvx_t$, which represent corrupted versions of the data ($\rvx_0$), hinders their ability to capture abstract semantic meaning, making them less suitable for applications requiring such representations \citep{diffae}. Additionally, the reliance on predefined additive Gaussian noise restricts them to continuous, fixed-length data, posing challenges for handling discrete sequences of varying lengths. Recent efforts \citep{li2022diffusionlm, genie, ardiffusion} adapt the approach to text modeling, but their performance still greatly lags behind pretrained autoregressive models.

\vspace{-5pt}
\section{Generalized Encoding-Decoding Diffusion Probabilistic Models (\modelnamenews)}
\label{sec:method}

We describe \modelnamenews, which integrate the core capabilities of generation, reconstruction, and compact representation within one framework. \modelnamenews thus combine the various advantages and applicabilities of 
different existing deep generative models while overcoming their limitations. 

\modelnamenews generalize the standard diffusion formulation described in \S\ref{sec:background}. The intuition is that the leading steps ($t=0, 1, \dots$) in the diffusion process can go beyond using only the common Gaussian noising-denoising. Instead, we can plug in more sophisticated ``noising'' and ``denoising'' operations. 
In particular, we use neural encoder and decoder with {\it learnable} parameters in place of ``noising'' and ``denoising'', respectively. The encoder allows the model to map its inputs into a low-dimensional space for a compact representation, and the decoder is to invert the encoding step and recover the inputs.
Crucially, this change is fully compatible with the diffusion objective 
in Eqs. (\ref{eq:elbo}) and (\ref{eq:ddpm_loss_kl}), 
allowing us to retain most of the critical derivations in the well-established formulation \citep{DBLP:conf/icml/Sohl-DicksteinW15,DBLP:conf/nips/HoJA20} and thus inherit the stable effective training of both the original denoising parameters and the new encoder-decoder parameters. 
As we discussed in \S\ref{sec:method:connections}, this differs crucially from latent diffusion models that train a VAE and a diffusion separately and are limited by the capability and quality of the VAE component. Our approach also shows superiority over previous joint training \citep{diffae,lsgm} which requires various approximations and can lead to unstable training and inferior performance.

\subsection{The Generalized Formulation}
\label{sec:method:our_diff}

We introduce learnable encoder-decoder to the leading $n$ steps in the diffusion process (i.e., for all steps $t$, where $1\leq t\leq n$). For simplicity and clarity, this work considers only the first step (i.e., $t=n=1$) in the diffusion process, though all the derivation can be applied to larger $n$ for $n$-layer hierarchical representations.

Let $\mathcal E_\lambda(\cdot)$ denote the neural encoder with free parameters $\lambda$. In the first diffusion step, the encoder transforms input $\rvx_0$ into a lower-dimensional latent vector $\rvx_1$, following:
\begin{equation}
\small
\begin{split}
  \qlambda(\rvx_1|\rvx_0) := \mathcal N(\rvx_1; \mathcal{E}_\lambda(\rvx_0), \beta_0 \mathbf{I})
\end{split}
\label{eq:our_forward}
\end{equation}
The diffusion process then continues by adding standard Gaussian noises to $\rvx_1$ step by step until reaching $\rvx_T$ as in \S\ref{sec:background}. Let $\mathcal{D}_\phi(\cdot)$ be the decoder with free parameters $\phi$. At the end of the denoising process, the decoder transforms $\rvx_1$ back to $\rvx_0$ in the data space. The actual form of $\mathcal{D}_\phi(\cdot)$ and its respective conditional $p_\phi(\rvx_0 | \rvx_1)$ can vary depending on the types of the data $\rvx_0$ (e.g., text, protein, image), as we detailed later. This offers added flexibility compared to the standard diffusion that is restricted to continuous data of fixed dimensionality.

As mentioned earlier, we can generalize the standard diffusion objective (Eqs. \ref{eq:elbo} and \ref{eq:ddpm_loss_kl}) to plug in the encoder and decoder seamlessly. The full derivations are provided in the appendix (\S\ref{app:derivation}). Briefly, we can adapt the left-hand side of \eqref{eq:ddpm_loss_kl} as:
\begin{equation}
\small
\begin{split}
   &\mathcal{L}(\lambda,\phi,\theta) \leq \\
   &\mathbb{E}_{q}\bigg[-\log p(\rvx_T) - \sum_{t=3}^{T}\log\frac{p_\theta(\rvx_{t-1}|\rvx_t)}{q(\rvx_{t-1}|\rvx_{t},\rvx_1)} \frac{q(\rvx_{t-1}|\rvx_1)}{q(\rvx_t|\rvx_1)}\bigg. \\
   &\bigg.\qquad\qquad\quad - \log \frac{p_\theta(\rvx_{1}|\rvx_2)}{q(\rvx_2|\rvx_{1})} 
   - \log \frac{p_\phi(\rvx_{0}|\rvx_1)}{q_\lambda(\rvx_1|\rvx_{0})} \bigg]. 
\end{split}
\label{eq:our_loss_kl_pre}
\end{equation}
That is, compared to the original \eqref{eq:ddpm_loss_kl}, we
replace the noising-denoising $\frac{p_\theta(\rvx_{0}|\rvx_1)}{q(\rvx_1|\rvx_{0})}$ in step $t=1$ with the parameterized encoder-decoder $\frac{p_\phi(\rvx_{0}|\rvx_1)}{q_\lambda(\rvx_1|\rvx_{0})}$; we also split from the sum an additional step $t=2$, which serves to form KL divergences after rearrangement (see \S\ref{app:deriv:our_kl} for more details), resulting in the final objective:
\begin{equation}
\small
\begin{split}
   &\mathbb{E}_{q} \left[ \underbrace{\text{KL}\left( q(\rvx_T|\rvx_1) || p(\rvx_T) \right)}_{\mathcal L_T} \right.\\
    &\qquad\quad + \sum_{t = 3}^T \underbrace{\text{KL}\left( q(\rvx_{t-1}|\rvx_t, \rvx_1) || p_\theta(\rvx_{t-1}|\rvx_t) \right)}_{\mathcal L_{t-1}} \\
    &\qquad\quad + \left. \underbrace{\text{KL}\left(\qlambda(\rvx_1|\rvx_0)||\ptheta(\rvx_1|\rvx_2)\right)}_{\mathcal L_{\text{align}}} -\underbrace{\log \pphi(\rvx_0|\rvx_1)}_{\mathcal L_{\text{rec}}}\right],
\end{split}
\label{eq:our_loss_kl}
\end{equation}
where $\mathcal{L}_T$ and $\mathcal{L}_{t-1}$ match the respective terms in \eqref{eq:ddpm_loss_kl}, and ${\mathcal L_{\text{align}}}$ and ${\mathcal L_{\text{rec}}}$ are new due to the generalization.

\paragraph{The ${\mathcal L_{\text{align}}}$ term}
This term serves to align the compact representations $\rvx_1$ from the encoder $q_\lambda(\rvx_1 | \rvx_0)$ and the denoising process $p_\theta(\rvx_1 | \rvx_2)$, resulting in a consistent latent representation space in the model. The KL divergence between the two Gaussians is written as:
\begin{equation}
\small
\begin{split}
    \mathcal{L}_{\text{align}} &= \bb E_q\left[ \gamma_1 \cdot\rho\|\mathcal E_\lambda(\rvx_0) - \rvmu_\theta(\rvx_2, 1)\|^2\right] + const,
\end{split}
\end{equation}

where $\rvmu_\theta$ is the Gaussian mean in \eqref{eq:reverse}; $\rho := \frac{\alpha_1(1-\balpha_1)}{\beta_1^2}$ and, following the approximation in \citet{DBLP:conf/nips/HoJA20}, $\gamma_1$ is set to $1$ in practice.

\paragraph{The ${\mathcal L_{\text{rec}}}$ term}
\label{sec:weight_w}
This terms is the data reconstruction loss that corresponds to $\mathcal{L}_0$ in standard diffusion objective \eqref{eq:ddpm_loss_kl}. However, thanks to the incorporation of the decoder $p_\phi(\rvx_0 | \rvx_1)$, we can generalize the original $\mathcal{L}_0$ (which is only for continuous data like image) to model various data modalities, such as text and protein sequences of varying lengths. For instance, for text data, the decoder can be a (pretrained) language model and the reconstruction loss becomes a standard sequential cross-entropy loss.

We provide more detailed derivations of the full final objective in \S\ref{app:deriv:our_final} and the complete training process in \S\ref{app:algorithm}. Thanks to the compatibility with the standard diffusion formulation \citep{DBLP:conf/nips/HoJA20}), we can largely follow their training recipes for effective training. In practice, we introduce a hyperparameter weight $w$ to balance the term $\mathcal{L}_{\text{rec}}$ against other terms like $\mathcal{L}_{t-1}$ and  $\mathcal{L}_{\text{align}}$, as detailed in \S\ref{app:deriv:our_final}.
In our experiments, using validation data, we set $w=8$ for text and $w=1$ for both protein and images.

\subsection{Generation, Reconstruction, and Representation}
\label{sec:method:3_func}

The trained \modelnamenews can naturally support the three core functionalities including generation, reconstruction, and representation, as well as the diverse applications built on top of the functionalities. To {\it generate} new samples, similar to standard diffusion, \modelnamenews simulate a random noise $\rvx_T$ from prior $p(\rvx_T)$, and go through the denoising process followed by decoding into a sample $\mathbf{\hat{x}}_0$ with the learned decoder. 
Due to the stochastic nature of the process (e.g., drawing random noise $\rvx_T$), \modelnamenews can generate highly diverse samples (\S\ref{sec:exp_text}). 
{\it Representing} an input with a compact vector is done straightforwardly by applying the learned encoder. To {\it reconstruct} the input, we further apply the decoder on the latent vector to obtain the reconstructed sample. The compact representation space also facilitates other tasks (Figure~\ref{fig:intro}), such as {\it interpolation} which is done by drawing an intermediate point between the representations of two given samples and decoding it into the data space, and {\it editing} which modifies the sample representation (e.g., with a latent classifier or latent vector arithmetic) followed by decoding. We disucss more details and analysis in experiments (\S\ref{sec:experiments}).

\subsection{Connections with Other Generative Models}
\label{sec:method:learnable-prior}\label{sec:method:connections}

We discuss the rich connections between \modelnamenews to other diverse deep generative models \citep{Goodfellow2016DeepGenerativeModels,hu2018unifying,Hu2022Toward}, providing insights into the integrated advantages of the new approach. \modelnamenews can be viewed as {\bf VAEs} with a learned diffusion model prior. This can be seen through \eqref{eq:our_loss_kl} where $\mathcal{L}_{\text{rec}}$ corresponds to the reconstruction loss in VAEs and $\mathcal{L}_{\text{align}}$ corresponds to the KL regularization with prior. In this perspective, the vanilla VAEs use a standard Gaussian prior $p(\rvx_1)$ while \modelnamenews learn the ``prior'' $p_\theta(\rvx_1 | \rvx_2)$ through diffusion modeling. Note that previous studies have also explored learnable priors in VAEs to overcome the difficulty of vanilla VAEs training and to improve expressiveness \citep{DBLP:journals/corr/DilokthanakulMG16,DBLP:conf/iclr/0022KSDDSSA17,DBLP:conf/aistats/TomczakW18,DBLP:conf/nips/RazaviOV19,DBLP:journals/corr/abs-2106-15671,lsgm,goyal2017nonparametric}. \citet{DBLP:journals/corr/abs-2106-15671} presented a preliminary study of diffusion priors in VAEs but without full derivations and experiments. If we replace all the noising-denoising steps in diffusion with parameterized encoders-decoders, we arrive at a {\bf hierarchical VAEs} \citep[e.g., NVAE, HiVAE, HVAE,][]{nvae, liu2023improved, hvae}.
Compared to hierarchical VAEs, \modelnamenews and diffusion allow for more effective training for generation thanks to the special parameterizations and training recipes \citep{DBLP:conf/nips/HoJA20}. 
{\bf Latent diffusion}~\citep{DBLP:conf/cvpr/EsserRO21} 
combines VAEs with diffusion but trains them separately. 
\modelnamenews, inherently with unified training,
offer a better semantic representation space than latent diffusion whose representation space is that of the vanilla VAEs. {\bf Diffusion Autoencoders} \citep[DiffAE,][]{diffae} and {\bf LSGM}~\citep{lsgm} explore joining training strategies of diffusion and autoencoder and introduce different approximation for tractability. 
\modelnamenews, benefiting from the well-established training objective and recipe from variational diffusion \citep{DBLP:conf/nips/HoJA20},
achieve more effective training and better performance (\S\ref{sec:exp_img}). Also, the applications of DiffAE and LSGM on text/protein data have not been explored. Compared to recent {\bf text diffusion models} \cite{yuan2023seqdiffuseq, he2022diffusionbert, genie, ye2023dinoiser} that generate text non-autoregressively, \modelnamenews can flexibly accommodate pretrained {\bf autogressive language models} as the decoder for much enhanced performance.

\section{Experiments}
\label{sec:experiments}

In this section, we present the main experimental results on text, image, and protein data. Additional results and related analysis can be found in \S\ref{app:experiments}.
\subsection{Text}
\label{sec:exp_text}

\paragraph{Setup} 
We use BERT-small~\citep{DBLP:conf/naacl/DevlinCLT19,bhargava2021generalization} as the encoder and GPT2-xl~\citep{radford2019language} as the decoder. 
A warmup training is done to align the encoder and decoder using the bookcorpus dataset~\citep{Zhu_2015_ICCV} without diffusion, followed by the proposed unified \modelnamenew training on the Yelp review dataset~\citep{shen2017style,li2018delete}.
We compare \modelnamenew with LatentOps~\citep{liu2022composable} and Optimus-DAAE~\citep{DBLP:conf/icml/ShenMBJ20,DBLP:conf/emnlp/LiGLPLZG20}, using the same architecture and training procedure. LatentOps is a form of latent diffusion model with successful applications on text data. It uses ordinary differential equations (ODEs), instead of discrete diffusion process as in standard latent diffusion models, on top of a separately trained VAE latent space. Optimus-DAAE is an improved variant of large-scale text VAEs \citep{DBLP:conf/emnlp/LiGLPLZG20} inspired by \citep{DBLP:conf/icml/ShenMBJ20}.
We also compare with latest text diffusion models, GENIE~\citep{genie} and AR-Diffusion~\cite{ardiffusion}. In addition, we compare with fine-tuned GPT2-xl and 20-shot GPT4~\cite{openai2023gpt4}. 
For comprehensive evaluation, we study tasks of generation, reconstruction, interpolation, and editing. More details are in \S\ref{app:exp_text:setup}.
We use BLEU score to measure \textbf{content} preservation, MAUVE~\citep{pillutla-etal:mauve:neurips2021} to assess the \textbf{distribution match} between the generated samples and the original data, and perplexity to evaluate text \textbf{fluency}. For text editing with latent arithmetic, transfer \textbf{accuracy} is measured as the geometric mean~\cite{liu2022composable} of the accuracy (by BERT classifiers) and the BLEU score (relative to the input text). 

\begin{table*}
\centering
\footnotesize
\begin{tabular}{lccccccc}
 \toprule
& Reconstruction& \multicolumn{3}{c}{Generation}& \multicolumn{1}{c}{Latent Arithmetic}&\multicolumn{2}{c}{Interpolation}\\\cmidrule(r){2-2}\cmidrule(r){3-5}\cmidrule(r){6-6}\cmidrule(r){7-8}
& Content$\uparrow$ & Fluency$\downarrow$ & Distr. Match$\uparrow$& Diversity$\uparrow$ & Accuracy$\uparrow$ & Fluency$\downarrow$  & Distr. Match$\uparrow$ \\\midrule
LatentOps& 87.6& 68.1& \uline{0.240}& 0.50& {\bf 57.3} & \uline{32.5}	&0.697 \\
Optimus-DAAE& 86.1& 94.1& 0.006& 0.17 & 51.0 & 33.7	& {\bf 0.770} \\
GENIE& 58.5& 337.6& 0.013&0.69 & 33.9& 258.2&0.029 \\
AR-Diffusion& 64.1& 157.8& 0.007&0.32 & 17.0& 163.9&0.012\\
GPT2& - & {\bf 15.0} & 0.015&0.65 & -&-&-\\
GPT4& {\bf 100} &25.7 &0.007 & {\bf0.87}& 49.3&39.7&0.010\\ \midrule
\modelnamenew & \uline{92.1}& \uline{ 16.4} & {\bf 0.977} &  \uline{0.79} & \uline{ 57.1} & {\bf 30.8} & \uline{0.763}\\\bottomrule             
\end{tabular}
\caption{\small Evaluation of Text Reconstruction, Generation (\S\ref{sec:exp_text:gen_con}), Latent Vector Arithmetic (\S\ref{sec:exp_text:transfer}), and Interpolation (\S\ref{sec:exp_text:interpolation}). The best results are highlighted in {\bf bold}, and the second-best results are \uline{underlined}.}
\label{tab:tab_rec_gen_tst_int}
\vspace{-10pt}
\end{table*}


\subsubsection{Text Generation and Reconstruction}
\label{sec:exp_text:gen_con}
Figure~\ref{fig:radar_text}(b) and Table~\ref{tab:tab_rec_gen_tst_int} show the results.
Autoencoding-based models (LatentOps and Optimus-DAAE) exhibit proficiency in reconstruction tasks, primarily because their objectives rely heavily on reconstruction quality. However, these models struggle to generate fluent text that aligns with the real data distribution. This limitation stems from the regularization in latent space designed to adhere to a specific prior distribution, such as a standard Gaussian. The regularization, while mathematically sound, often results in a significant disparity between the model's latent space and the actual latent distribution observed in real-world scenarios \citep{yang2017improved,DBLP:conf/emnlp/LiGLPLZG20}. 
The text diffusion models GENIE and AR-Diffusion
apply a diffusion process over the token space. Despite the reconstruction-based training objective, these models fall short of both reconstruction and generation. The results highlight the difficulty of standard diffusion modeling for discrete text data.
The fine-tuned GPT2 autoregressive model excels in generation fluency but
performs poorly in terms of generation diversity and domain match. It also fails for reconstruction. 
Despite the general capability of GPT4, it fails to capture the domain-specific text characteristics via in-context demonstrations. 
In comparison, \modelnamenew achieves strong performance across all metrics for both reconstruction and generation.
The learned autoencoding ensures effective reconstruction, while the diffusion process introduces dynamic regularization to the latent space, ensuring superior quality in text generation. 

\subsubsection{Sentence Interpolation}
\label{sec:exp_text:interpolation}
In this experiment, we randomly selected 200 samples from the test set, dividing them into two groups of 100 each. We performed interpolation between these groups, encoding the first 100 samples as $\rvx_1^1$ and the second 100 as $\rvx_1^2$. Given that both sets of latent distributions adhere to a Gaussian distribution, we utilized spherical linear interpolation (Slerp) as per Shoemake's method~\citep{DBLP:conf/siggraph/Shoemake85}, employing the formula $\text{Slerp}(\rvx_1^1,\rvx_1^2;\alpha)$ where $0 \leq \alpha \leq 1$. Our aim was to generate a series of sentences that smoothly transition in semantics, achieving a fluent blend.

We set the number of interpolation steps to 10 and evaluated the quality at each step to gauge performance. Particularly, the interpolation results at the midpoint are shown in Table~\ref{tab:tab_rec_gen_tst_int}. \modelnamenew demonstrates a consistent ability to produce fluent sentences that closely align with the original data distribution. In contrast, baseline models such as GENIE and AR-Diffusion struggled, primarily due to their lack of a semantically coherent latent space for whole sentences. Additionally, we assessed the outcomes using MAUVE and BLEU metrics, with detailed interpolation examples provided in \S\ref{app:exp_text:interpolation}.

\subsubsection{Latent Vector Arithmetic}
\label{sec:exp_text:transfer}
Text editing (e.g., changing the text sentiment) necessitates proficiency in all three fundamental abilities. Previous research \citep{DBLP:conf/icml/ShenMBJ20,hu2017toward} demonstrated that sentence representations can capture linguistic relationships through simple arithmetic operations. We use this approach as a proxy for evaluating the quality of the learned latent space.
Specifically, we use the sentiment attribute and evaluate how well the learned latent space can support the inference of text sentiment change with simple arithmetic operations in the latent space.
We first obtain a positive-sentiment latent vector by averaging the latent vectors of 100 positive sentences randomly sampled from the dataset. A negative-sentiment latent vector is obtained similarly with 100 negative sentences.
We then compute the difference between the two sentiment vectors, denoted as $\rvv$. 
Given a new sentence, it is first encoded to obtain its latent vector, $\rvx_1 = \mathcal{E}_\lambda(\rvx_0)$. 
The sentiment-transferred sentence is then acquired via $\mathcal{D}_\phi(\rvx_1 \pm k\rvv)$, where $k$ serves as a weight modulating the degree of transfer. We tried different $k\in\text{range}(1,5,.5)$ and show the best one in Table~\ref{fig:radar_text}. preservation score. 
\modelnamenew is on par with LatentOps and outperforms others, validating a well-structured latent space.

\subsubsection{Training Efficiency}
\label{sec:train_efficiency}
\begin{figure}[t]
    \centering
    \includegraphics[width=0.45\textwidth]{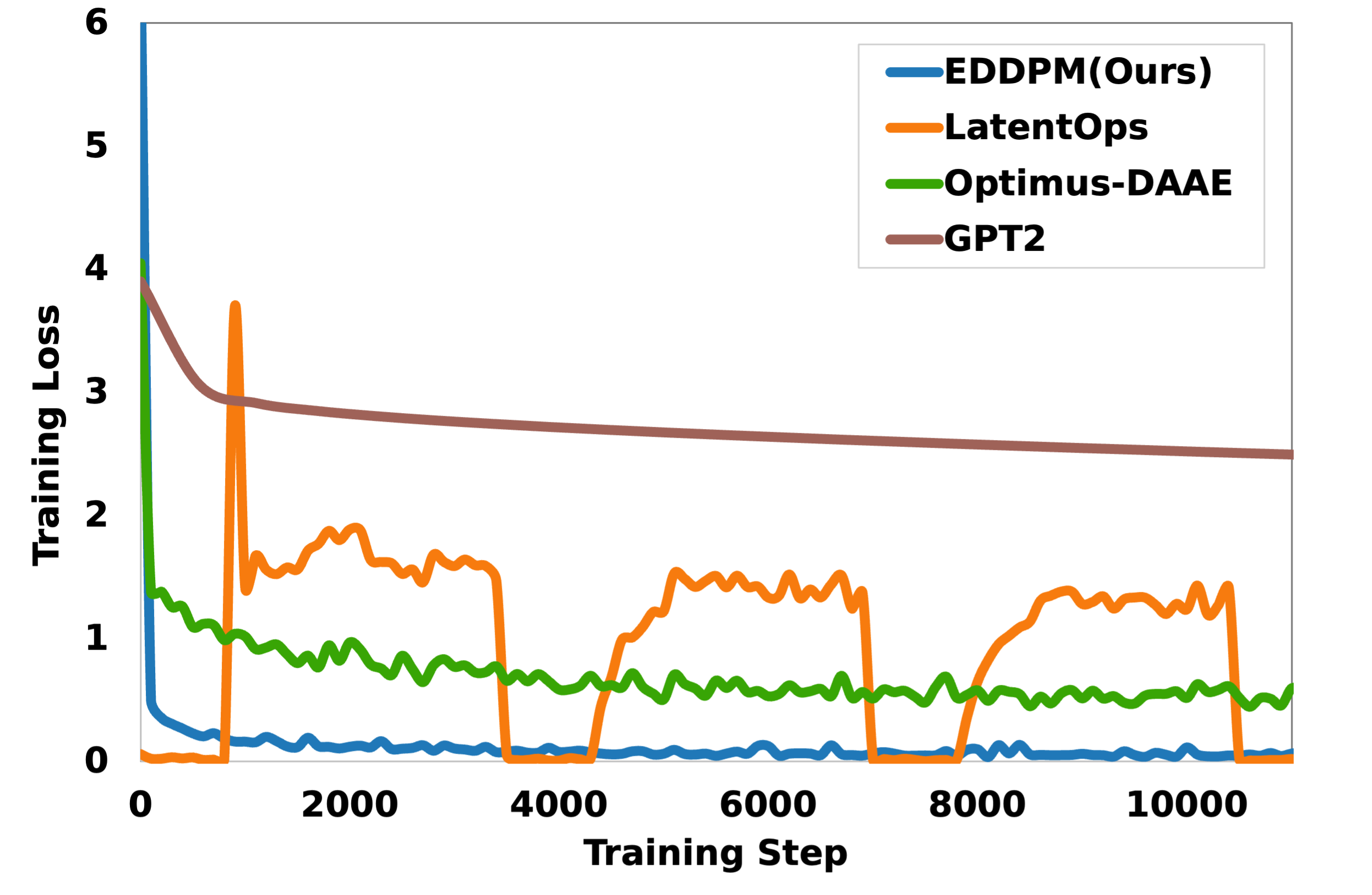}  
    \vspace{-10pt}
    \caption{
    Training curves. \modelnamenew converges fast in a stable way. LatentOps, which depends on text VAE, relies on periodic scheduling of the weight of the prior regularization. The scheduling is common for training text VAE models \citep{DBLP:conf/emnlp/LiGLPLZG20}, and leads to the ups-and-downs in the loss curve, indicating a difficult trade-off between generation and reconstruction capabilities.
    }
    \label{fig:training_curves}
\end{figure}

To show the training efficiency of \modelnamenews, we measured the training cost on text data. The results shown in Table~\ref{tab:text_efficiency} validate that training \modelnamenews is as efficient as training VAEs and variants (e.g., Optimus-DAAE, LatentOps) with the same encoder-decoder architecture. Due to the added diffusion process, the training time of \modelnamenew for each epoch is ~1.3x that of LatentOps and Optimus-DAAE. However, thanks to the stable training process inherited from the well-established DDPM formulation, \modelnamenews avoid the different training tricks necessary in VAEs and variants, such as beta-annealing~\cite{DBLP:conf/conll/BowmanVVDJB16}, free bits~\cite{kingma2017improving}, and cyclic annealing schedule~\cite{fu2019cyclical}. This allows \modelnamenews to converge in fewer steps, leading to similar overall training cost. Figure~\ref{fig:training_curves} shows the training curves.
\begin{table}[t]
\small
    \centering
    \footnotesize
    \begin{tabular}{lcc}\toprule
      \multicolumn{1}{l}{Models}   & Time / Epoch   & Overall Training Time \\\midrule
LatentOps	&13.98 mins	&6.98 hrs \\
Optimus-DAAE&14.49 mins	&7.02 hrs\\\midrule
\modelnamenew	&19.35 mins	&7.41 hrs\\\bottomrule
    \end{tabular}
    \vspace{-5pt}
    \caption{Training cost.}
    \label{tab:text_efficiency}
\end{table}

\subsection{Image}
\label{sec:exp_img}

\paragraph{Setup}

We adopt UNet \citep{UNet} as the encoder and the diffusion-based model in DiffAE \citep{diffae}) as the decoder. 
Following DiffAE, we train our model and then evaluate the reconstruction and generation abilities on the popular FFHQ~\citep{StyleGAN}, CelebA~\citep{ProgressiveGAN}, LSUN-Bedroom, and LSUN-Horse~\citep{lsun} datasets. We use the CelebA-HQ dataset to perform image manipulation tasks.
In addition to DiffAE, we compare our model with a number of latest models, including the Latent Diffusion Model \citep[LDM,][]{Rombach2021HighResolutionIS}, DDIM~\citep{DDIM}, StyleGAN-XL~\citep{stylegan-xl}, NVAE~\citep{nvae}, and Consistency Models~\citep{cm}. 
We compare \modelnamenew with these models where existing corresponding model checkpoints exist.  
Following the common practice, we use FID and reconstruction-FID (rFID) to evaluate the generation quality and reconstruction quality, respectively.
To systematically
evaluate \modelnamenew, besides the individual evaluations of reconstruction and generation, we also perform image \textit{interpolation} and \textit{manipulation} (\S\ref{sec:exp_img:manipulation}) tasks, which can reflect the integrated ability. 
We perform interpolation within the latent space and then reconstruct images from the resulting interpolated representations. Given two latent vectors, \( \rvx_1^1 \) and \( \rvx_1^2 \), we employ linear interpolation using the formula: \( \alpha \rvx_1^1 + (1-\alpha) \rvx_1^2 \), where \( \alpha \) represents the interpolation ratio and we use $\alpha=0.2$ and $0.4$. This interpolated latent vector is subsequently fed into the decoder to produce the interpolated image. In total, 50k images are used for interpolation and the obtained images are measured against 50k images in the training set to obtain FID score. 
The detailed image experimental setup is in \S\ref{app:exp_img:setup}.

\subsubsection{Generation, Reconstruct., Interpolation}
\begin{figure}[t]
  \centering
  \includegraphics[width=1.0\linewidth]{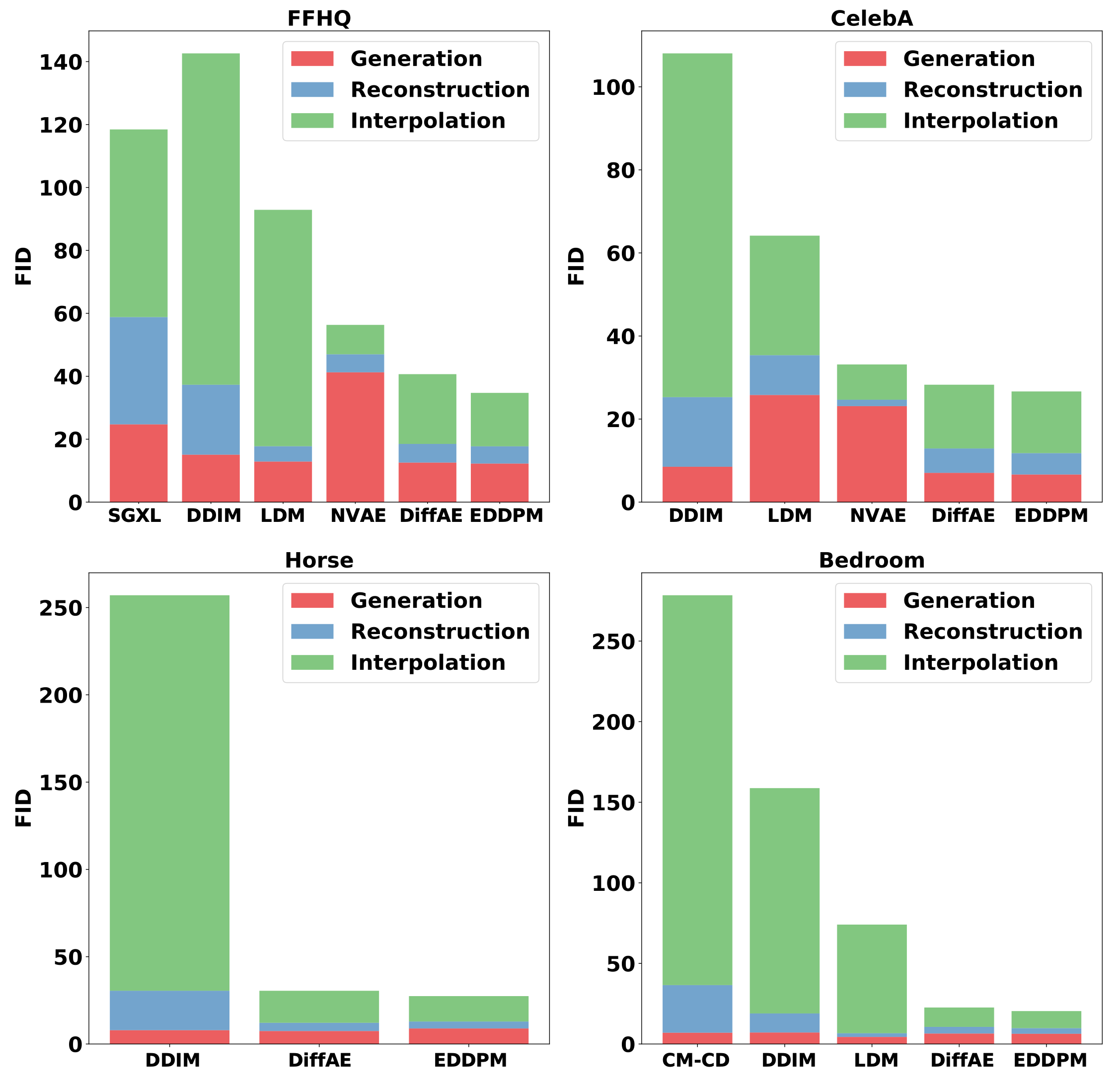}
  \vspace{-.5cm}
  \caption{Overall performance comparison on images. SGXL stands for StyleGAN-XL. We stack the FIDs of generation, reconstruction, and interpolation together to show the overall performances of different models.}
  \label{fig:image_results}
    \vspace{-0.5cm}
\end{figure}

The overall performances for generation, reconstruction and interpolation are shown in Figure \ref{fig:image_results}. For all experiments requiring diffusion process, we fix the inference steps to be $T=50$. From the figure, 
We could observe: 
(1) \modelnamenew consistently achieves superior or at least comparable performance across all three evaluated tasks, and is the best in terms of aggregate performance across the datasets examined; (2) NVAE demonstrates robust capabilities in the tasks of reconstruction and interpolation. However, its performance in the generative task is notably less competitive compared with other models; (3) LDM, while demonstrating commendable performance in reconstruction, falls short in maintaining quality during interpolation. The detailed experimental results are shown in Table \ref{tab:overall_performance_comparison} (\S\ref{ssub:experimental_results}).

\begin{figure}[t]
    \centering    
    \includegraphics[width=\linewidth]{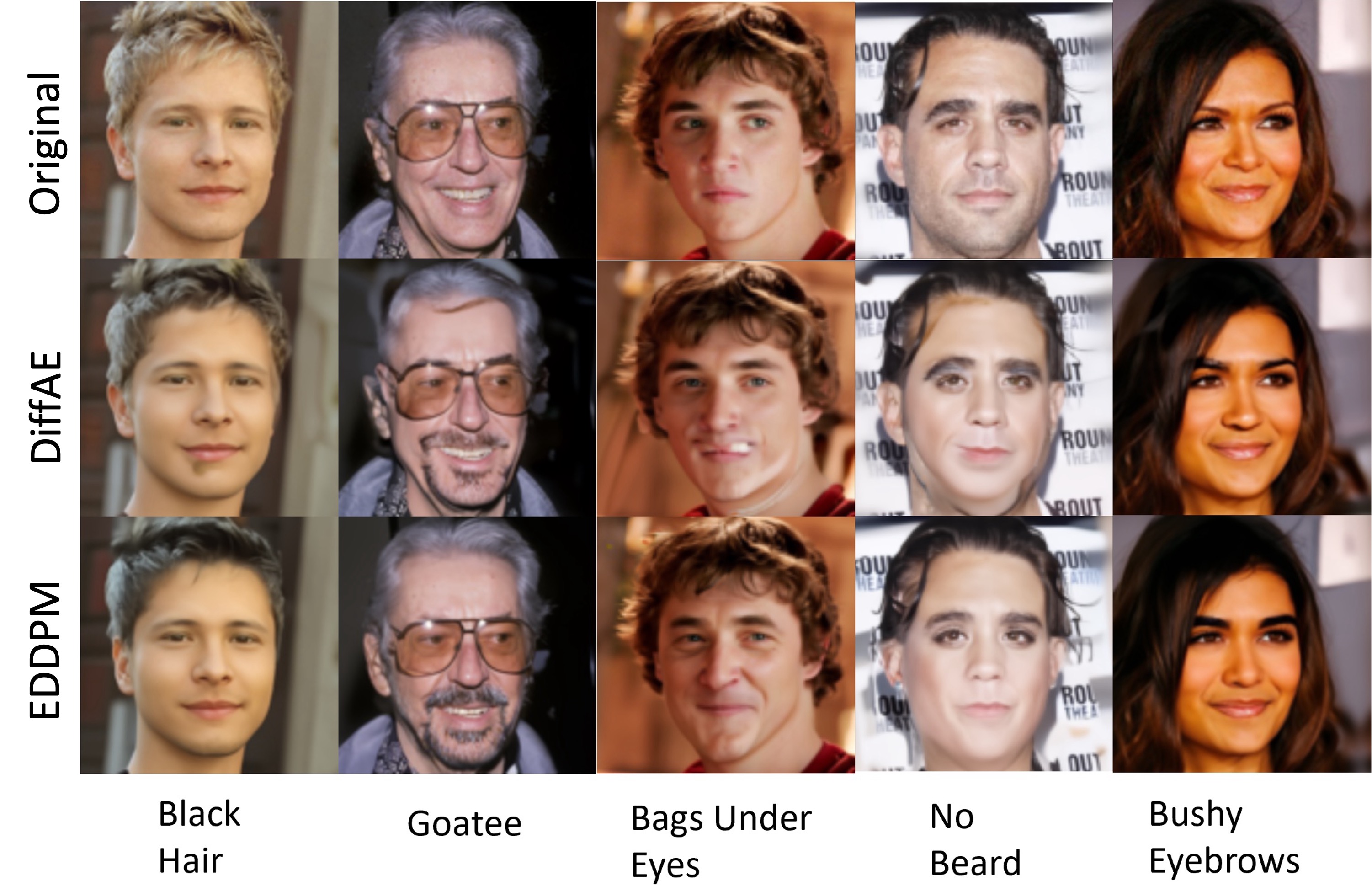}
    \vspace{-0.7cm}
    \caption{\textbf{Image Manipulation:} The procedure for manipulation is detailed in Section \ref{sec:exp_img:manipulation}. The class names provided at the bottom indicate the target class towards which the images are being manipulated.}
    \vspace{-10pt}
    \label{fig:manipulation_main}
\end{figure}

\subsubsection{Image Manipulation}
\label{sec:exp_img:manipulation}

We deploy our model trained on FFHQ to the CelebA-HQ domain in a zero-shot fashion. 
We select the CelebA-HQ dataset for this task due to the availability of 40 binary classification category labels. 

\begin{wraptable}{r}{4cm} 
\footnotesize
\centering
\hspace*{-8pt}
\vspace{-8pt}
\resizebox{4cm}{!}{
    \begin{tabular}{c|cc}
    \toprule
        Model & $\epsilon=0.1$ & $\epsilon=0.3$ \\
    \midrule
        DiffAE & 19.62 & 23.69 \\
        \modelnamenew & \textbf{13.48} & \textbf{18.43} \\
    \bottomrule
    \end{tabular}}
\caption{Manipulation (FID).}
\label{tab:overall_comparison_on_manipulation}
\vspace{-8pt}
\end{wraptable}
Following DiffAE, we train a linear classifier, \( \mathbf{y} = \mathbf{w}^\top \mathbf{x}_1 + b \), on 70\% of the training data for each attribute. 
This classifier predicts the attribute based on the representation $\rvx_1$. To construct the manipulated image representation, we use \( \mathbf{x}_1' = \mathbf{x}_1 + \epsilon\mathbf{w} \), where \( \epsilon \) is a scalar determining the manipulation magnitude. The manipulated representation \( \mathbf{x}_1' \) is then used to generate the corresponding image. 
We show the examples in Figure \ref{fig:manipulation_main}. We compare ours with DiffAE as we follow its setting and it is also one of the state-of-the-art attribute-based image manipulation methods. 
Our evaluation focuses on two aspects: 1) {\it image quality post-manipulation}: as shown in Table~\ref{tab:overall_comparison_on_manipulation}, \modelnamenew consistently yields high-quality images after manipulation. 2) {\it alignment with target class}: we test the linear classifier on the remaining 30\% of the dataset. In terms of weighted AUC, \modelnamenew's representation achieves 0.915, closely matching DiffAE's 0.917. Notably, \modelnamenew surpasses in accuracy, registering 0.893 against DiffAE's 0.795. 
A detailed AUC comparison can be found in \S\ref{app:exp_img}.

\subsection{Protein Sequences}
\label{sec:exp_protein}

\paragraph{Setup}
Following the setup of ReLSO \citep{castro2022relso}, a protein autoencoder, we adopt a simple transformer as the encoder and convolutional layers as the decoder. 
In line with the settings of ReLSO, we jointly train a simple regressor to predict the fitness values from the latent embeddings of protein sequences. 
This fitness value, representing some desired properties of proteins, serves as a performance metric with higher values indicating superiority.
Our models are trained and evaluated on the Gifford \citep{10.1093/bioinformatics/btz895} dataset and GFP \citep{gfp} dataset separately. 
In addition to ReLSO, we provided quantitative comparisons between our models and several baseline models, namely NOS \citep{gruver2023protein} which utilized transformer-based discrete diffusion and Gaussian diffusion model for protein design, as well as vanilla VAEs \citep{kingma2013auto}. Detailed experimental setups are in \S\ref{app:exp_prot:setup}. We evaluate \modelnamenew's representation ability in \S\ref{sec:exp_protein:rep} and its generation ability through both protein optimization (\S\ref{sec:exp_protein:opt}) and protein generation (\S\ref{app:exp_prot:gen}). The evaluation on \modelnamenew's reconstruction ability is provided in the appendix at \S\ref{app:exp_prot:rec}.

 \begin{figure}
  \centering
  \includegraphics[width=0.45\textwidth]{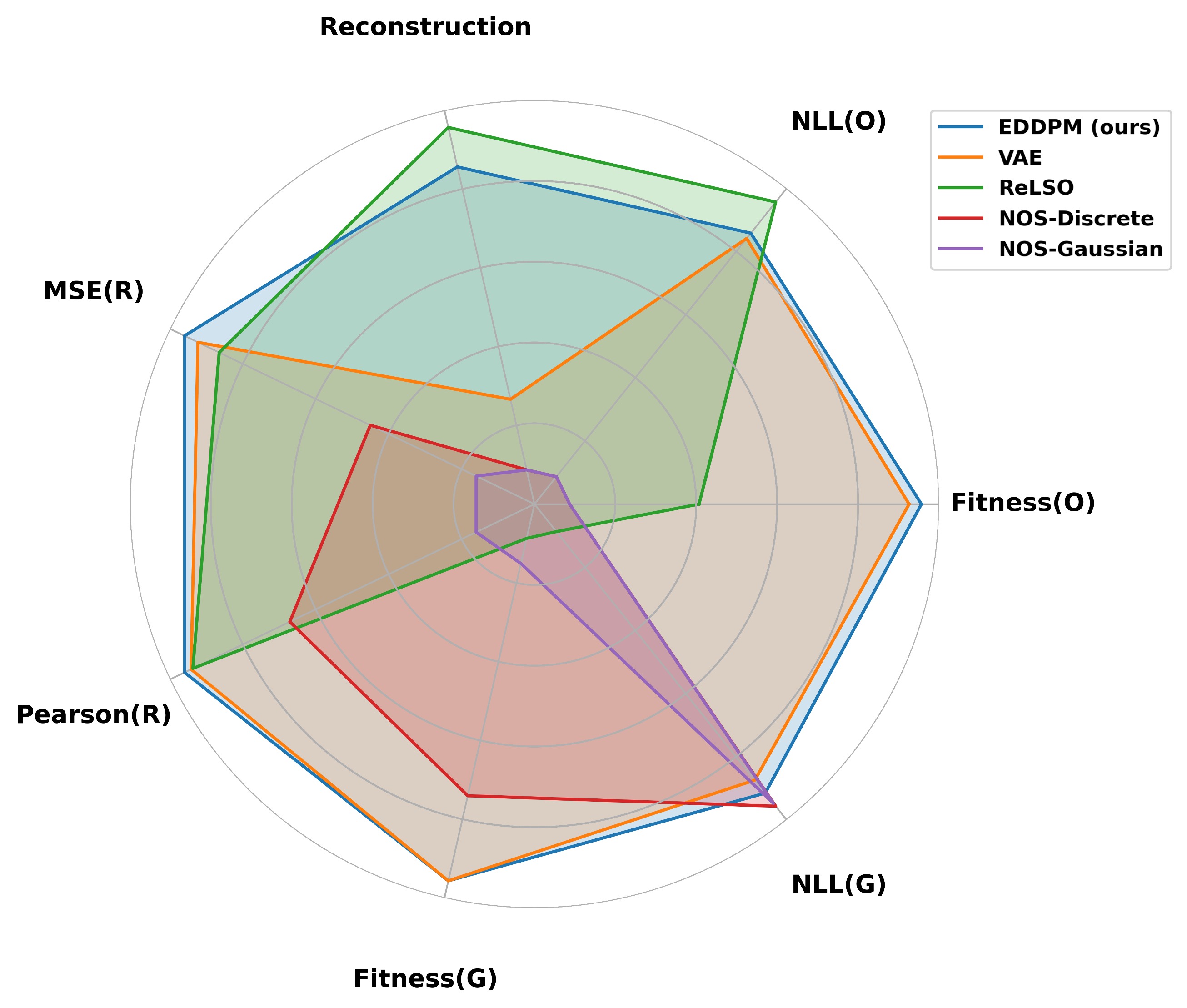}
  \vspace{-0.2cm}
  \caption{\small{\modelnamenew shows robust performance on fundamental tasks about protein sequences. R, G, and O denote \textbf{R}epresentation, \textbf{G}eneration, and \textbf{O}ptimization} respectively. Results are normalized for better visualization. Detailed results in \S\ref{app:exp_prot}}
  \label{fig:radar_protein}
\end{figure}

\subsubsection{Protein Representation}
\label{sec:exp_protein:rep}
\begin{figure}
  \centering
  \includegraphics[width=1.1\linewidth]{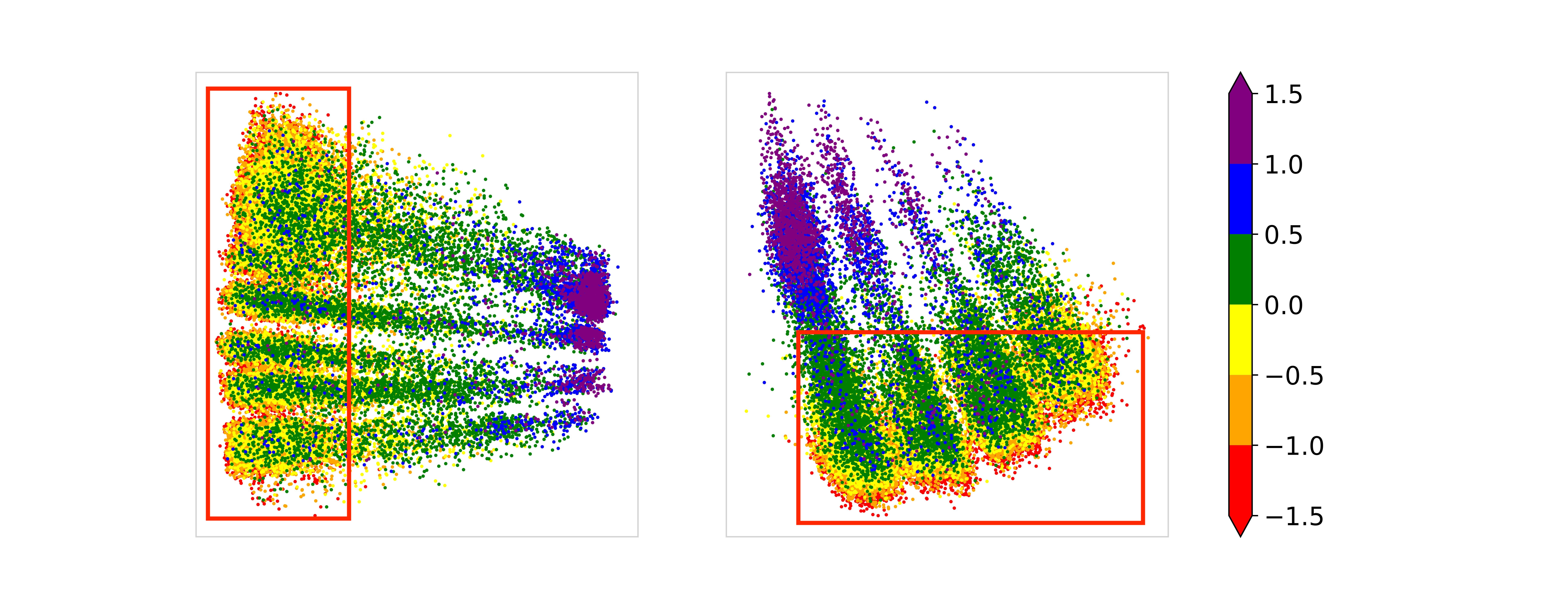}
  \vspace{-1cm}
  \caption{\small{Protein Latent Space of ReLSO (left) and \modelnamenew (right). Different colors represent different fitness values of the corresponding protein sequence.}}
    \vspace{-0.4cm}
  \label{fig:protein_graphs}
\end{figure}

After training, each protein sequence in the test set is transformed into a latent representation, upon which the fitness value is predicted using the regressor in the latent space. Evaluation metrics, including MSE, L1 Norm, Pearson and Spearman correlation coefficients, are computed between the predicted values and the ground truth. The results are presented in Table~\ref{tab:protein_representation}.  The regressor trained in the latent space of \modelnamenew demonstrates superior performance over the regressors from the baseline models on all four metrics on the Gifford Dataset. This suggests that \modelnamenew obtained more refined representations for the protein sequences, leading to more accurate predictions by the regressor. This is also evident from the visualization of the latent space. In Figure~\ref{fig:protein_graphs}, we display the latent spaces for both ReLSO (left) and \modelnamenew (right), coloring proteins by their respective fitness value intervals. In the latent space of ReLSO in Figure~\ref{fig:protein_graphs}(left), 
proteins with fitness values less than $0$ exhibits an overlap (inside the red box). While some overlap persists across intervals inside the red box in \modelnamenew's latent space in Figure~\ref{fig:protein_graphs} (right), 
the delineation between each interval is much clearer.

\subsubsection{Protein Optimization}
\label{sec:exp_protein:opt}
\begin{figure}
  \centering
  \includegraphics[width=1.1\linewidth]{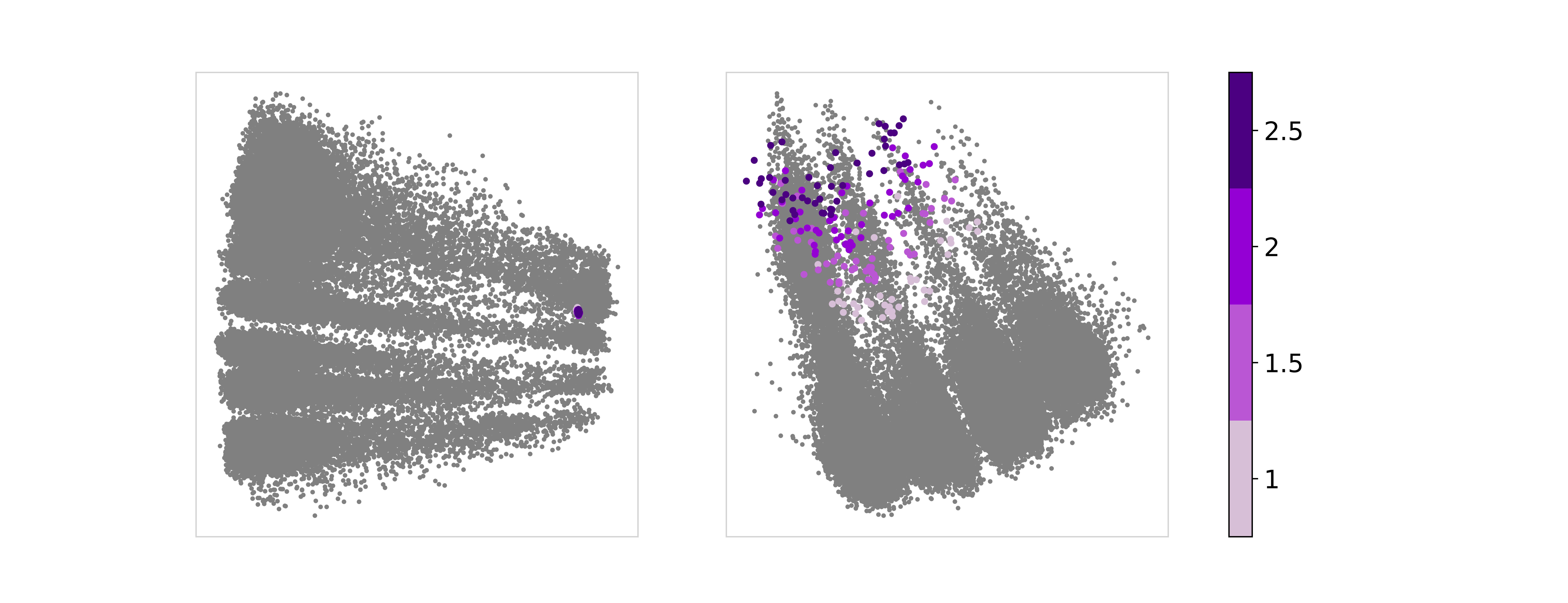}
  \vspace{-1cm}
  \caption{Optimized Protein Sequence of ReLSO (left) and \modelnamenew (right) in the latent space.}
    \label{fig:protein_optimization}
\end{figure}

We optimize a protein sequence by optimizing its corresponding representation in the latent space. 
We adopt the sampling algorithm introduced in LatentOps \citep{liu2022composable} that solves an ODE involving the regressor \citep{liu2022composable}. This approach requires a target fitness value to guide the optimization. We set this value to 1, 1.5, 2, and 2.5 for Gifford proteins; 3, 4 for GFP proteins, all of which represent reasonably high fitness values within the respective dataset. 
Visualizations of the optimized sequence with ReLSO and \modelnamenew algorithms are shown in Figure~\ref{fig:protein_optimization}. The grey dots represent proteins from the Gifford dataset, while colored dots represents optimized proteins with varying target fitness values. As shown in Figure \ref{fig:protein_optimization} (left),
the pseudo-convex nature of ReLSO leads to convergence of optimized sequences to a singular point, revealing a lack of diversity.
In contrast, as depicted in Figure \ref{fig:protein_optimization} (right),
our model not only achieves superior fitness values but also fosters a broader protein variety.
For a detailed quantitative analysis, refer to \S\ref{app:exp_prot:opt}.

\section{Other Related Works}
\label{sec:related_work}
{To tackle the limitations of current generative models, past research has ventured into developing hybrid models by combining VAEs ~\citep{DBLP:conf/nips/RazaviOV19,DBLP:conf/nips/VahdatK20,dai2018diagnosing} and GANs ~\citep{DBLP:conf/cvpr/KarrasLAHLA20,DBLP:conf/cvpr/EsserRO21,DBLP:journals/pami/XiaZYXZY23} to create VAE-GAN hybrids ~\citep{DBLP:conf/icml/LarsenSLW16,hu2018unifying,wu2020improving,DBLP:journals/corr/abs-2101-02908}. Recently, diffusion models ~\citep{DBLP:conf/icml/Sohl-DicksteinW15,DBLP:conf/nips/HoJA20} have been noted for their generative ability, yet they fall short in providing a robust semantic latent space. Efforts like ~\citet{DBLP:journals/corr/abs-2106-15671} and ~\citet{lsgm} explored the parameterization of the variational prior with diffusion models.
Latent diffusion models (e.g., stable diffusion) ~\citep{Rombach2021HighResolutionIS} merge autoencoder and diffusion model architectures, optimizing diffusion training with the VAE's pivotal role in image dimension compression. A deep dive can be found in \S\ref{app:LDM}. Despite these advancements, it remains elusive to achieve all three capabilities of representation, generation, and reconstruction effectively. \modelnamenews aim to tackle these challenges. }

\section{Conclusion}
This work generalized the diffusion model to introduce \modelnamenews, an innovative generative framework designed to seamlessly integrate the three core functionalities for generative models: generation, reconstruction and representation. To this end, we incorporated parameterized encoder and decoder transformations into the conventional diffusion process. We derived an end-to-end training objective from the data likelihood inspired from the widely-used DDPM training framework.
Experimental results across image, text, and protein sequence data demonstrate that \modelnamenews consistently outperform strong baselines across all three core functionalities. 
Building a compact high-quality representation space with the versatile utilities (generation, reconstruction, encoding, interpolation, editing) can be significant for building new foundation models, such as world models \citep{hu2023language,lecun2022path,assran2023self}, for more robust machine reasoning \citep{hao2023reasoning,wong2023word}.

\paragraph{Limitations}
In this work, \modelnamenews have not been verified on other data modalities such as videos, audios, and time series. It would be interesting to see how \modelnamenews will perform on all the diverse forms of data, and even multi-modal data (such as text-image). In addition, we have only parameterized the first diffusion step with learnable encoding/decoding. A single latent vector representation could be limited to capture relevant information of data for different tasks. We would like to investigate our approach with more encoding/decoding-based diffusion steps for  hierarchical latent representations.

\section*{Impact Statement}
There are potential societal consequences associated with diffusion modeling and deep generative models in general regarding content creation, privacy, and others.

\bibliography{icml2024}
\bibliographystyle{icml2024}

\newpage
\appendix
\onecolumn
\onecolumn

\section{Derivation}
\label{app:derivation}
\subsection{Derivation of our ELBO loss}
\label{app:deriv:our_elbo}
Below is a derivation of the error bound of the log-likelihood loss.
\begin{align}
\small
   &\mathbb{E}_{\rvx_0\sim q(\rvx_0)}\left[-\log p_{\phi,\theta}(\rvx_0)\right] \\
   &\leq \mathbb{E}_{\rvx_0\sim q(\rvx_0)}\left[-\log p_{\phi,\theta}(\rvx_0) + \text{KL}(q_\lambda(\rvx_{1:T}|\rvx_0)||p_{\phi,\theta}(\rvx_{1:T}|\rvx_0))\right] \\
   &=  \mathbb{E}_{\rvx_0\sim q(\rvx_0)}\left[-\log p_{\phi,\theta}(\rvx_0) +  \mathbb{E}_{\rvx_{1:T}\sim q_\lambda(\rvx_{1:T}|\rvx_0)}[\log\frac{q_\lambda(\rvx_{1:T}|\rvx_0)}{p_{\phi,\theta}(\rvx_{1:T}|\rvx_0)}]\right] \\
   &= \mathbb{E}_{\rvx_0\sim q(\rvx_0)}\left[-\log p_{\phi,\theta}(\rvx_0) +  \mathbb{E}_{\rvx_{1:T}\sim q_\lambda(\rvx_{1:T}|\rvx_0)}[\log\frac{q_\lambda(\rvx_{1:T}|\rvx_0)}{p_{\phi,\theta}(\rvx_{0:T})/p_{\phi,\theta}(\rvx_0)}]\right] \\
   &=  \mathbb{E}_{\rvx_0\sim q(\rvx_0)}\left[-\log p_{\phi,\theta}(\rvx_0) +  \mathbb{E}_{\rvx_{1:T}\sim q_\lambda(\rvx_{1:T}|\rvx_0)} [\log\frac{q_\lambda(\rvx_{1:T}|\rvx_0)}{p_{\phi,\theta}(\rvx_{0:T})}]+\log p_{\phi,\theta}(\rvx_0)\right]\\
   &= \mathbb{E}_{\rvx_0\sim q(\rvx_0),\rvx_{1:T}\sim q_\lambda(\rvx_{1:T}|\rvx_0)}\left[\log\frac{q_\lambda(\rvx_{1:T}|\rvx_0)}{p_{\phi,\theta}(\rvx_{0:T})}\right] \\
   &= \mathbb{E}_{q}\left[ -\log\frac{p_{\phi,\theta}(\rvx_{0:T})}{q_\lambda(\rvx_{1:T}|\rvx_0)} \right]
\end{align}

\subsection{Derivation of our KL loss}
\label{app:deriv:our_kl}
Below is a derivation of \eqref{eq:our_loss_kl}.
\begin{align}
   \mathcal{L}(\lambda,\phi,\theta) =&\mathbb{E}_q\left[ -\log\frac{p_{\phi,\theta}(\rvx_{0:T})}{q_\lambda(\rvx_{1:T}|\rvx_0)} \right] \\ 
   &= \mathbb{E}_q\left[ -\log\frac{p_\phi(\rvx_0|\rvx_1)p_\theta(\rvx_{1:T})}{q(\rvx_{2:T}|\rvx_1)q_\lambda(\rvx_1|\rvx_0)} \right]\\
   &= \mathbb{E}_q\left[ -\log\frac{p_\phi(\rvx_0|\rvx_1)p(\rvx_T) \prod_{t=1}^{T}p_\theta(\rvx_{t-1}|\rvx_t)}{\prod_{t=1}^{T}q(\rvx_t|\rvx_{t-1})q_\lambda(\rvx_1|\rvx_0)} \right]\\
   &= \mathbb{E}_q\left[-\log p(\rvx_T) - \sum_{t=3}^T\log\frac{p_\theta(\rvx_{t-1}|\rvx_{t})}{q(\ttm)}  - \log\frac{\ptheta (\rvx_1|\rvx_2)}{q(\rvx_2|\rvx_1)}
   - \log\frac{\pphi(\rvx_0|\rvx_1)}{\qlambda(\rvx_1|\rvx_0)} \right] \\  %
   &= \mathbb{E}_q\left[-\log p(\rvx_T) - \sum_{t=3}^T\log\frac{\ptheta(\rvx_{t-1}|\rvx_{t})}{q(\tmtz)}\cdot \frac{q(\tmz)}{q(\tz)}  - \log\frac{\ptheta (\rvx_1|\rvx_2)}{q(\rvx_2|\rvx_1)}
   - \log\frac{\pphi(\rvx_0|\rvx_1)}{\qlambda(\rvx_1|\rvx_0)} \right] \\
   &= \mathbb{E}_q\left[-\log \frac{p(\rvx_T)}{q(\rvx_T|\rvx_1)} - \sum_{t=3}^T\log\frac{\ptheta(\rvx_{t-1}|\rvx_{t})}{q(\tmtz)} - \log\frac{\ptheta (\rvx_1|\rvx_2)}{\qlambda(\rvx_1|\rvx_0)} - \log\pphi(\rvx_0|\rvx_1) \right] \\
   &= \mathbb{E}_{q} \left[ \underbrace{\text{KL}\left( q(\rvx_T|\rvx_1) || p(\rvx_T) \right)}_{\mathcal L_T} +\right. \sum_{t = 3}^T \underbrace{\text{KL}\left( q(\rvx_{t-1}|\rvx_t, \rvx_1) || p_\theta(\rvx_{t-1}|\rvx_t) \right)}_{\mathcal L_{t-1}} \notag \\
   &\qquad\qquad\qquad\left. +\underbrace{\text{KL}\left(\qlambda(\rvx_1|\rvx_0)||\ptheta(\rvx_1|\rvx_2)\right)}_{\mathcal L_{\text{align}}} -\underbrace{\log \pphi(\rvx_0|\rvx_1)}_{\mathcal L_{\text{rec}}}\right]. \label{eq:app_our_objective}
\end{align}

\subsection{Derivation of each KL term}
\label{app:deriv:our_kl_each}
In line with \cite{DBLP:conf/nips/HoJA20}, our objective exclusively involves KL divergences between Gaussians, thus facilitating closed-form evaluations.

\paragraph{For \(\mathcal{L}_T\)}: The posterior distribution \( q \) lacks learnable parameters due to the deterministic forward mapping from \( \rvx_1 \) to \( \rvx_T \). Specifically, we have \( q(\rvx_T|\rvx_1) = \gN(\rvx_T; \sqrt{\balpha_T} \rvx_1, (1-\balpha_T) \rmI) \). When \( \balpha_T \approx 1 \), this simplifies to \( q(\rvx_T|\rvx_1) = \gN(\rvx_T; \mathbf{0}, \rmI) \). Given this property, \( \mathcal{L}_T \) remains constant during training and can be excluded from optimization.

\paragraph{For \( \mathcal{L}_{1:T-1} \)}: This term aligns with the conventional diffusion model. 
Given the two conditional Gaussian distributions, 
$q(\rvx_{t-1}|\rvx_t,\rvx_1)$ is derived as
\begin{align}
    q(\rvx_{t-1}|\rvx_t,\rvx_1)=\gN(\rvx_{t-1};\Tilde{\rvmu}(\rvx_t,\rvx_1),\Tilde\beta_t), \qquad\qquad\qquad\qquad\qquad \\
    \text{where}\quad \Tilde\rvmu_t(\rvx_t, \rvx_1) \defeq \frac{\sqrt{\bar\alpha_{t-1}}\beta_t }{1-\bar\alpha_t}\rvx_1 + \frac{\sqrt{\alpha_t}(1- \bar\alpha_{t-1})}{1-\bar\alpha_t} \rvx_t \quad \text{and} \quad
\tilde\beta_t \defeq \frac{1-\bar\alpha_{t-1}}{1-\bar\alpha_t}\beta_t,  \label{eq:q_posterior_mean_var}
\end{align}
$\mathcal L_{t-1}$ can be compactly represented as:
\begin{align}
  \gL_{t-1}
   = \bb E_q\left[ \frac{1}{2\sigma_t^2} \|\tilde\rvmu_t(\rvx_t,\rvx_1) - \rvmu_\theta(\rvx_t, t)\|^2 \right] + C, \label{eq:mu_squared}
\end{align}
where \(C\) is a constant, independent with \(\theta\). We can expand \eqref{eq:mu_squared} further by reparameterizing \(q(\rvx_t|\rvx_1)\) as \(\rvx_t(\rvx_1,\rveps)=\sqrt{\balpha_t}\rvx_1+\sqrt{1-\balpha_t}\rveps\) for \(\rveps\sim\gN(\mathbf 0,\mathbf I)\) and reparemetrizing $\rvmu_\theta(\rvx_t, t)$ as $\frac{1}{\sqrt{\bar{\alpha}_t}}(\rvx_t - \sqrt{1-\bar\alpha_t} \rveps_\theta(\rvx_t, t))$:
\begin{align}
    \gL_{t-1} = \mathbb{E}_{\rvx_1\sim q(\rvx_1),\rveps\sim\mathcal N(\mathbf 0, \mathbf I)}\left[\gamma_t\left\| \boldsymbol\epsilon-\boldsymbol\epsilon_\theta\left(\sqrt{\bar\alpha_t}\rvx_1+\sqrt{1-\bar\alpha_t}\boldsymbol\epsilon,t\right) \right\|^2 \right],
     \label{eq:gamma_t}
\end{align}
where $\gamma_t=\frac{\beta_t^2}{2\sigma_t^2 \alpha_t (1-\bar\alpha_t)}$.

\paragraph{For \( \mathcal{L}_{\text{align}} \)}: This loss serves to align the latent representations derived from both the encoder and the diffusion model, ensuring consistent latent spaces across the framework. 
Both \(p_\theta(\rvx_1|\rvx_2)\) and \(q_\lambda(\rvx_1|\rvx_0)\) are Gaussian distributions. 
Therefore, the KL divergence has the similar form as \eqref{eq:mu_squared}, while the \(\tilde\rvmu_t(\rvx_t,\rvx_1)\) becomes the mean of \(q_\lambda(\rvx_1|\rvx_0)\), which corresponds to \(\mathcal E_\lambda(\rvx_0)\). So the loss function becomes:
\begin{align}
     \mathcal{L}_{\text{align}} 
     &= \bb E_q\left[ \frac{1}{2\sigma_1^2} \|\mathcal E_\lambda(\rvx_0) - \rvmu_\theta(\rvx_2, 1)\|^2 \right] \\
     &= \bb E_q\left[ \frac{\beta_1^2}{2\sigma_1^2 \alpha_t (1-\bar\alpha_1)}\frac{\alpha_1(1-\balpha_1)}{\beta_1^2} \|\mathcal E_\lambda(\rvx_0) - \rvmu_\theta(\rvx_2, 1)\|^2\right] \\
     &=\bb E_q\left[ \gamma_1 \cdot\rho\|\mathcal E_\lambda(\rvx_0) - \rvmu_\theta(\rvx_2, 1)\|^2\right] 
\end{align}
where $\gamma_1$ is consistent with \eqref{eq:gamma_t}, $\rho:=\frac{\alpha_1(1-\balpha_1)}{\beta_1^2}$ is a constant, and $\rvmu_\theta(\rvx_2, 1)$ is reparemeterized under the same reparemetrization trick used in $\mathcal{L}_{1:T-1}$, i.e., \(\rvmu_\theta(\rvx_2, 1) = \frac{1}{\sqrt{\bar{\alpha}_1}}(\rvx_2 - \sqrt{1-\bar\alpha_1} \rveps_\theta(\rvx_2, 1))\).

\paragraph{For \(L_{\text{rec}}\)}: This is the reconstruction loss corresponding to the one in VAE.
Depending on the data type, it can be formulated using different loss functions. In our model, we employ MSE loss for continuous data like images and cross-entropy for discrete data, including texts and proteins.

\subsection{Derivation of our Final loss}
\label{app:deriv:our_final}
The final objective in \eqref{eq:our_loss_kl} in  can be reformulated as:

\begin{align}
     &\mathbb{E}_{q,\rveps}\left[\sum_{t=3}^{T}    \gamma_t\left\| \boldsymbol\epsilon-\boldsymbol\epsilon_\theta\left(\rvx_t,t\right) \right\|^2 
    + \gamma_1 \cdot\rho\|\mathcal E_\lambda(\rvx_0) - \rvmu_\theta(\rvx_2, 1)\|^2 + \mathcal L_{\text{rec}}
    \right] \\
    &=\mathbb{E}_{q,\rveps}\left[\sum_{t=3}^{T}    \gamma_t\left\| \boldsymbol\epsilon-\boldsymbol\epsilon_\theta\left(\rvx_t,t\right) \right\|^2 
    + \gamma_1 \cdot\rho\|\mathcal E_\lambda(\rvx_0) - \rvmu_\theta(\rvx_2, 1)\|^2 + T\cdot\frac{1}{T}\mathcal L_{\text{rec}}
    \right] \\
    &=\mathbb{E}_{q,\rveps}\left[\sum_{t=3}^{T}   \gamma_t\left( \left\| \boldsymbol\epsilon-\boldsymbol\epsilon_\theta\left(\rvx_t,t\right) \right\|^2+\frac{1}{\gamma_t T}\mathcal L_{\text{rec}} \right)
    +\gamma_1\left( \rho\|\mathcal E_\lambda(\rvx_0) - \rvmu_\theta(\rvx_2, 1)\|^2 + \frac{1}{\gamma_1 T}\mathcal L_{\text{rec}}\right)
    \right] 
\end{align}
We further introduce a hyperparameter $w$ to balance the diffusion/align loss with the reconstruction loss, which gives us the following loss:
\begin{align}
\label{eq:detail_final_loss}
    \mathcal L^{\text{final}} = \mathbb{E}_{q,\rveps}\left[\sum_{t=3}^{T}   \underbrace{\gamma_t\left( w \left\| \boldsymbol\epsilon-\boldsymbol\epsilon_\theta\left(\rvx_t,t\right) \right\|^2+\frac{1}{\gamma_t T}\mathcal L_{\text{rec}} \right)}_{\mathcal L^{\text{final}}_{t-1}}
    +\underbrace{\gamma_1\left( w \left(\rho\|\mathcal E_\lambda(\rvx_0) - \rvmu_\theta(\rvx_2, 1)\|^2\right) + \frac{1}{\gamma_1 T}\mathcal L_{\text{rec}}\right)}_{\mathcal L^{\text{final}}_1}
    \right] .
\end{align}

\subsection{Derivation of our loss from a learnable prior perspective}
\label{app:deriv:our_vae}
The following is a derivation of our loss from the VAE perspective as shown in \S\ref{sec:method:learnable-prior}, 
the objective from a learnable prior perspective.
\begin{align}
    \mathcal{L}(\lambda, \phi, \theta; \rvx_0) 
    &= \mathbb{E}_q\left[ -\log\frac{p_{\phi,\theta}(\rvx_{0:T})}{q_\lambda(\rvx_{1:T}|\rvx_0)} \right] \\
    &=\mathbb{E}_q\left[ -\log\frac{p_\phi(\rvx_0|\rvx_1)p_{\theta}(\rvx_{1:T})}{q_\lambda(\rvx_1|\rvx_0)q(\rvx_{2:T}|\rvx_1)} \right] \\
    &= \mathbb{E}_q\left[ -\log\frac{p_\phi(\rvx_0|\rvx_1)p_{\theta}(\rvx_1)}{q_\lambda(\rvx_1|\rvx_0)} \right] \\
    &= -\mathbb{E}_{q_\lambda(\rvx_1 | \rvx_0)}[\log p_\phi(\rvx_0| \rvx_1)] + \text{KL}(q_\lambda(\rvx_1 | \rvx_0) || p_\theta(\rvx_1))\label{eq:our_vae_deriv}
\end{align}

\section{Algorithm}
Below shows the complete training algorithm of \modelnamenews.
\label{app:algorithm}
\begin{algorithm}
\caption{Training}\label{alg:training}
\begin{algorithmic}[1]
\Repeat
    \State $\mathbf{x} \sim q(\mathbf{x})$
    \State $\rveps_0 \sim \mathcal{N}(0, \mathbf{I})$
    \State $\rvx_1 = \mathcal{E}_\lambda(\rvx_0) + \beta_0 \rveps_0$
    \State $t \sim \text{Uniform}(\{1, \ldots, T-1\})$
    \State $\rveps \sim \mathcal{N}(0, \mathbf{I})$
    \State $\rvx_t = \sqrt{\bar{\alpha}_t} \rvx_1 + \sqrt{1 - \bar{\alpha}_t} \rveps$
    \If{t == 1}
        \State $\rvmu_\theta(\rvx_2, 1) = \frac{1}{\sqrt{\bar{\alpha}_1}}(\rvx_2 - \sqrt{1-\bar\alpha_1} \rveps_\theta(\rvx_2, 1))$
        \State Take gradient descent step on $\mathcal L^{\text{final}}_1$
    \Else
        \State Take gradient descent step on $\mathcal L^{\text{final}}_t$
    \EndIf
\Until{converged}
\end{algorithmic}
\end{algorithm}

\section{Details of Experiments}
\label{app:experiments}
This section is the counterpart of the experiment section \S\ref{sec:experiments}.

\subsection{Text}
\label{app:exp_text}
\subsubsection{Setup}
\label{app:exp_text:setup}
\paragraph{Model Architecture}
We closely adhere to the experimental setup presented in LatentOps~\citep{liu2022composable}. For our encoder, denoted as $\mathcal E_\lambda$, we utilize the BERT-small model\footnote{The BERT model follows the Apache 2.0 License.}\citep{DBLP:conf/naacl/DevlinCLT19,bhargava2021generalization}. As for the decoder, represented as $\mathcal D_\phi$, we employ the GPT2-xl architecture\footnote{The GPT2 model follows the MIT License.}\citep{radford2019language}. Our diffusion model is constructed using a straightforward MLP with skip connections, as inspired by~\citep{diffae}. The latent dimension is set to 128.

Building upon the methodologies of \citet{DBLP:conf/emnlp/LiGLPLZG20,liu2022composable}, 
we equip the pretrained language model (LM) with a linear layer that precedes the LM, facilitating the passage of $\rvx_1$ to the decoder. To maintain generative capabilities and acclimate the LM to the latent space, we incorporate an additional transformer layer between the original first layer and the embedding layer of the LM, fostering adaptability. During the training phase, our optimization is confined to the MLP layers, the embedding layer, the newly inserted transformer layer, the encoder, and the diffusion model with all other parameters remaining frozen. 

\paragraph{Dataset}
Regarding our dataset selection, we commence with the bookcorpus dataset~\citep{Zhu_2015_ICCV} to train the autoencoder in the absence of the diffusion model. Subsequently, we engage in joint training of the model with diffusion, utilizing the Yelp review dataset\footnote{The datasets are distributed under CC BY-SA 4.0 license.}~\citep{shen2017style}, which has been preprocessed by \citet{li2018delete}. It's noteworthy that Yelp serves as a sentiment dataset, encompassing approximately 179K negative and 268K positive sentences.

\paragraph{Baselines}
To ensure a comprehensive and fair comparison, our approach is primarily benchmarked against two key models: LatentOps~\citep{liu2022composable} and Optimus-DAAE~\citep{DBLP:conf/icml/ShenMBJ20,DBLP:conf/emnlp/LiGLPLZG20}. We have maintained consistency in both architecture and training procedures across these comparisons. The primary difference lies in the training objectives. Optimus-DAAE represents a modified, large-scale autoencoder. It enhances text generation capabilities by reconstructing sentences from their slightly altered versions and integrates adversarial training with a denoising objective.

In addition, we have conducted comparisons with the latest advancements in text diffusion models, namely GENIE~\citep{genie} and AR-Diffusion~\cite{ardiffusion}. GENIE stands out as a non-autoregressive, large-scale pre-training text diffusion framework designed for text generation. AR-Diffusion, on the other hand, is an auto-regressive text diffusion model characterized by its multi-level diffusion strategy, which operates at both sentence and token levels. To evaluate these models comprehensively, we trained both GENIE and AR-Diffusion using the Yelp dataset and applied them to various downstream tasks. In addition to the text diffusion models, our comparative analysis extends to include two state-of-the-art purely auto-regressive models: GPT2 and GPT4~\cite{openai2023gpt4}. Specifically, we adapted GPT2 for our experiments by fine-tuning it on the Yelp dataset; however, its application was limited to text generation tasks due to its model architecture and capabilities. Conversely, GPT4 was evaluated across a broader range of tasks, utilizing a diverse set of prompts designed to explore its extensive generative potential and versatility.

\paragraph{Tasks}
In our study, we aim to comprehensively assess the holistic performance of \modelnamenews by evaluating it across three foundational capabilities: generation, reconstruction, and latent vector arithmetic. To achieve this, we focus on four specific tasks: generation, reconstruction, interpolation, and text style transfer using latent vector arithmetic. Each of these tasks is selected for the following reasons:
\begin{itemize}
\item \textbf{Good Reconstruction:} A key aspect of our model's performance is its ability to accurately preserve content. This is particularly evident in its reconstruction capability, which is critical for tasks like text style transfer and interpolation, where maintaining the original content's integrity is essential.
\item \textbf{Robust Representation:} Our model aims to do more than just preserve content; it strives to capture the intrinsic meaning, nuances, and essential features of the input text. This ability is vital for tasks requiring a deep understanding of the source material, such as text style transfer.

\item \textbf{Fluent Generation:} The ultimate test of our model's effectiveness lies in the quality of its output. For \modelnamenew, the fluency and coherence of the generated text are crucial indicators of its robust generative capabilities. This is demonstrated in generation tasks, text style transfer, and interpolation.
\end{itemize}

By evaluating \modelnamenew on these four tasks, we can effectively reflect and analyze its performance in terms of the three core functionalities.

\subsubsection{Detailed Results}
\label{app:exp_text:interpolation}
We present detailed results of our text experiments. The results for reconstruction, generation, and style transfer are displayed in Table~\ref{tab:tab_rec_gen_tst}. Additionally, results of interpolation are separately provided in two tables: Table~\ref{tab:tab_inter_content} focuses on content preservation, while Table~\ref{tab:tab_inter_fluency} addresses fluency and divergence.
 
\begin{table}[H]
\centering
\small
\begin{tabular}{lcccccc}
 \toprule
& Reconstruction& \multicolumn{2}{c}{Generation}& \graylineempty{Style Transfer (Arithmetic)}\\\cmidrule(r){2-2}\cmidrule(r){3-4}\cmidrule(r){5-7}
& Content$\uparrow$ & Fluency$\downarrow$ & Divergence$\uparrow$ & Accuracy$\uparrow$ & Attribute$\uparrow$  & Content$\uparrow$ \\\midrule
LatentOps~\cite{liu2022composable}& 87.6& 68.1& 0.240& 57.3& 71.5&45.9 \\
Optimus-DAAE~\cite{DBLP:conf/icml/ShenMBJ20}& 86.1& 94.1& 0.006& 51.0 & 74.7&34.8\\
GENIE~\cite{genie}& 58.5& 337.6& 0.013& 33.9& 20.8&55.2 \\
AR-Diffusion~\cite{ardiffusion}& 64.1& 157.8& 0.007& 17.0& 4.6& 62.7\\
GPT2~\cite{radford2019language}& - &15.04 & 0.015& -&-&-\\
GPT4~\cite{openai2023gpt4}& 100 &25.65 &0.007 & 49.3&80.5&30.1\\ \midrule
\modelnamenew & 92.1& 16.4& 0.977& 57.1 & 77.6& 42.1\\\bottomrule             
\end{tabular}
\caption{Evaluation of Text Reconstruction, Generation and Style Transfer. Accuracy of style transfer is the geometric mean of attribute score (classifier) and content score (BLEU).}
\label{tab:tab_rec_gen_tst}
\end{table}

\begin{table}[H]
\centering
\small
\begin{tabular}{lcccccc}
 \toprule
\multicolumn{1}{c}{Models}              & \multicolumn{1}{c}{$\alpha=0.0/1.0$} & \multicolumn{1}{c}{$\alpha=0.1/0.9$} & \multicolumn{1}{c}{$\alpha=0.2/0.8$} & \multicolumn{1}{c}{$\alpha=0.3/0.7$} & $\alpha=0.4/0.6$  & $\alpha=0.5$   \\ \midrule
{LatentOps~\cite{liu2022composable}}    & 0.0 / 92.0& 0.0 / 90.1& 0.0 / 83.7& 0.8 / 63.2& 0.9 / 35.2 & 9.5  \\
{Optimus-DAAE~\cite{DBLP:conf/icml/ShenMBJ20}}& 0.0 / 91.4& 0.0 / 87.2& 0.0 / 73.5& 0.5 / 46.5& 2.2 / 21.0 & 7.8  \\
{GENIE~\cite{genie}}& 0.8 / 66.2& 0.6 / 67.8& 0.8 / 65.6& 0.8 / 66.6& 0.5 / 52.1 & 10.1 \\
{AR-Diffusion~\cite{ardiffusion}} & 0.7 / 60.9& 0.9 / 87.3& 0.9 / 86.9 & 1.0 / 84.8& 0.6 / 62.2 & 15.9 \\\midrule
{\modelnamenew}& 0.0 / {97.2}& 0.0 / {95.6}&0.3/ {91.7}& 0.4 / 69.9& 3.0 / 30.3 & 11.5\\\bottomrule
\end{tabular}
\caption{Evaluation of Text Interpolation Content Consistency. This table quantifies the consistency of interpolated text between two input sentences using BLEU scores. The degree of interpolation is denoted by $\alpha$. A higher BLEU score indicates greater text consistency with the corresponding input sentence.}
\label{tab:tab_inter_content}
\end{table}
\begin{table}[H]
\centering
\small
\begin{tabular}{lcccccc}
 \toprule
\multicolumn{1}{c}{Models}              & \multicolumn{1}{c}{$\alpha=0.0/1.0$} & \multicolumn{1}{c}{$\alpha=0.1/0.9$} & \multicolumn{1}{c}{$\alpha=0.2/0.8$} & \multicolumn{1}{c}{$\alpha=0.3/0.7$} & $\alpha=0.4/0.6$  & $\alpha=0.5$   \\ \midrule
{LatentOps~\cite{liu2022composable}}    & 19.5 / \textbf{0.871}& 19.5 / \textbf{0.886}& 19.7 / 0.867& 21.3 / 0.798& 26.9 / 0.856& 32.5 / 0.697\\
{Optimus-DAAE~\cite{DBLP:conf/icml/ShenMBJ20}}& 20.2 / 0.848& 20.4 / 0.847& 20.9 / 0.849& 26.0 / 0.799& 31.6 / 0.655& 33.7 / 0.623 \\
{GENIE~\cite{genie}}& 49.8 / 0.013& 45.4 / 0.010& 46.9 / 0.015 & 45.0 / 0.008& 77.4 / 0.027& 300.0 / 0.029 \\
{AR-Diffusion~\cite{ardiffusion}} & 63.1 / 0.008& 25.9 / 0.007& 26.3 / 0.006& 29.1 / 0.008& 60.4 / 0.011& 156.0 / 0.012\\
GPT4\cite{openai2023gpt4} & -&-&-&-&-&39.7 / 0.010 \\\midrule 
{\modelnamenew}& \textbf{17.8} / 0.867& \textbf{17.9} / 0.884& \textbf{18.1} / \textbf{0.868}& \textbf{19.9} / \textbf{0.910}& \textbf{26.5} / \textbf{0.909}& \textbf{30.8} /\textbf{0.763} \\\bottomrule
\end{tabular}
\caption{Evaluation of Text Interpolation Fluency and Divergence. The table presents perplexity scores / MAUVE scores. Lower perplexity scores signify higher fluency and higher MAUVE scores represent smaller gap with the training data.}
\label{tab:tab_inter_fluency}
\end{table}

\subsection{Image}
\label{app:exp_img}
\subsubsection{Setup}
\label{app:exp_img:setup}

\paragraph{Model Architecture} 
In line with the architecture presented by Diffusion Autoencoders (DiffAE) \citep{diffae}, our model is structured with an encoder designed as a UNet and a decoder functioning as a conditional diffusion-based model at the pixel level. Given the latent semantic representation \( \rvx_1 \) and a random Gaussian sample \( \mathbf{x}_T \) that shares the same dimensionality as the raw data \( \mathbf{x} \), the decoder employs reverse diffusion transitions to produce the output \( \hat{\mathbf{x}} \). Complementing this, we integrate an additional standard diffusion process, with transitions realized through a straightforward MLP fortified with skip connections.

\paragraph{Dataset} 
Following the approach of DiffAE, we train our model and subsequently evaluate its reconstruction and generation capabilities on FFHQ~\citep{StyleGAN}, CelebA~\citep{ProgressiveGAN}, LSUN-Bedroom and LSUN-Horses~\citep{lsun}. To assess the representation ability, we employ the CelebA-HQ dataset~\citep{ProgressiveGAN}.

\paragraph{Hyperparameter settings} We trained our model on two Nvidia A100-SXM4-40GB GPUs with a batch size of 100. For evaluation purposes, we sampled 50,000 images to compute the FID, setting total steps \( T = 100 \) for both the diffusion process and the decoder at every 500,000 training steps interval. The optimization was carried out using the Adam Optimizer, with a learning rate of \( 1 \times 10^{-4} \) and no weight decay. The image dimensions inputted into the model were consistently set at \( 128 \times 128 \) for FFHQ and \( 64 \times 64 \) for CelebA.

\paragraph{Evaluation Metrics}
For the assessment of the generated images' quality, we resort to the Fréchet Inception Distance (FID) \citep{FID}, a widely-accepted metric in the field. To assess the fidelity of our reconstruction, we employ the reconstruction-FID (rFID) metric.

\paragraph{Baselines}
In our experiments, we compare our method against the following prominent baselines:
\begin{enumerate}
    \item \textbf{DDIM}~\citep{DDIM}: Following the implementation by \cite{diffae}, we ensure a fair comparison by adopting the DDIM approach described therein.
    
    \item \textbf{DiffAE}~\citep{diffae}: In this baseline, the encoder-decoder structure is trained in isolation, separate from the latent diffusion model.

    \item \textbf{StyleGAN-XL}~\citep{stylegan-xl}: StyleGAN-XL is the scaled version of StyleGAN3 generator on Imagenet, achieving state-of-the-art results on large-scale image synthesis. The provided checkpoint on FFHQ is of resolution 256, thus we perform all the experiments with $256\times 256$ but resize the images into $128\times 128$ for FID calculation. 

    \item \textbf{NVAE}~\citep{nvae}: Nouveau VAE (NVAE) is a deep hierarchical VAE built for image generation. 

    \item \textbf{Consistency Models}: Consistency Models (CM) are proposed to overcome the limitation that diffusion models are too slow during generation. It could achieve new state-of-the-art results with one-step generation. In our experiments, we adopt the best setting which is run by two steps. Specifically, we use the model \texttt{\url{openai/diffusers-cd_bedroom256_lpips}} in our experiments as it performs better than \texttt{\url{openai/diffusers-cd_bedroom256_l2}} (Generation FID of the former is lower).  Similar to StyleGAN-XL, we conduct all the experiments in the resolution $256\times 256$ and calculate FID with resolution $128\times 128$. 
    
    \item \textbf{Latent Diffusion Model (LDM)}~\citep{Rombach2021HighResolutionIS}: For this method, the Variational AutoEncoder (VAE) is pretrained prior to training the latent diffusion model. We leverage the pre-trained weights from \cite{Rombach2021HighResolutionIS} to generate images with a resolution of $224 \times 224$. Subsequent to generation, these images are downscaled to a $64 \times 64$ resolution for Fréchet Inception Distance (FID) computation. For tasks including reconstruction, interpolation, and manipulation, the VAE from the LDM approach serves as the benchmark.
\end{enumerate}

\subsubsection{Detailed Experimental Results}
\label{ssub:experimental_results}
\paragraph{Detailed Overall Performance Comparison for Generation, Reconstruction, and Interpolation.} The results are shown in Table \ref{tab:overall_performance_comparison}. Note that we draw Figure \ref{fig:image_results} according to the generation, reconstruction performances, and interpolation results with $\alpha=0.4$. {As for the results of CM-CD, the reported FID for generation in the original paper~\citep{cm} is 5.22, whereas we obtained an FID of 7.01. This discrepancy arises due to two main reasons: (1) We calculate the reference statistics (mean and variance) using the original images, resulting in slightly different statistics compared to those provided here\footnote{\url{https://github.com/openai/guided-diffusion/tree/main/evaluations}}. (2) The original model was trained on images with a resolution of 256x256. Consequently, we need to convert the generated 256x256 images to 128x128 for evaluation, which may introduce differences in FID. When using the pre-calculated statistics and evaluating the model at a resolution of 256x256, we obtain an FID of 5.82, which is close to the 5.22 reported in the original paper. Using the newly calculated statistics, the FID is 6.56. Upon converting the images to 128x128, the FID becomes 7.01.}

\begin{table*}[t]
\centering
\resizebox{\textwidth}{!}{
    \begin{tabular}{c|c|ccc|ccc|ccc|ccc}
    \toprule
        \multirow{2}{*}{Dataset} & \multirow{2}{*}{Model} & \multicolumn{3}{c|}{Generation FID} & \multicolumn{3}{c|}{Reconstruction rFID} & \multicolumn{3}{c|}{$\alpha=0.2$} & \multicolumn{3}{c}{ $\alpha=0.4$} \\
        & & T=10 & T=20 & T=50 & T=10 & T=20 & T=50 & T=10 & T=20 & T=50 & T=10 & T=20 & T=50  \\
    \midrule
        \multirow{6}{*}{FFHQ 128} & LDM & 67.78 & 30.43 & 12.90 & \grayline{\underline{4.87}} & \grayline{21.29} & \graylineempty{75.13} \\
        & NVAE & 
        \grayline{41.26}
        & \grayline{5.73} &
        \grayline{\underline{6.28}} & \graylineempty{\underline{9.34}}\\
        & StyleGAN-XL & \grayline{24.72} & \grayline{34.09}  & \grayline{46.82} & \graylineempty{59.64} \\
        & DDIM & 29.56 & 21.45 & 15.08 & 88.22 & 45.30 & 22.23 & 144.40 & 104.91 & 75.81 & 181.07 & 131.80 & 105.31 \\
         & DiffAE & 20.80	& 16.70 &	12.57 & 12.59 &	9.23 &	5.93 & 13.25 & 11.33 & 9.38 & 22.93 & 23.01 & 22.17\\
         & \modelnamenew &  \textbf{18.41} &	\textbf{14.38} & \textbf{12.26} & 
         \textbf{11.50} & \textbf{8.17} & 
         \textbf{5.48} & \textbf{12.57} & \textbf{9.80} & \textbf{6.66} &  \textbf{19.11} & \textbf{18.36} & \textbf{16.98}\\
    \midrule
        \multirow{5}{*}{CelebA 64} & LDM & 41.87 & 31.40 & 25.80 & \grayline{9.58} & \grayline{9.95} & \graylineempty{28.78} \\
        & NVAE &  \grayline{23.11} & \grayline{\underline{1.55}} & \grayline{\underline{3.03}} & \graylineempty{\underline{8.49}} \\ 
        & DDIM & 16.38 & 12.70 & 8.52 & 78.44 & 20.20 & 16.76 & 148.22 & 80.06 & 51.84 & 163.26 & 103.01 & 82.77 \\
         & DiffAE & 12.92 & 10.18 & 7.05 & 14.14 & 10.09 & 5.87 & 11.09 & 9.30 & 6.90 & 17.13 & 16.82 & 15.35 \\
         & \modelnamenew & \textbf{12.35 } & \textbf{9.49} & \textbf{6.65} & \textbf{11.84} & \textbf{8.60} & \textbf{5.15} & \textbf{9.75} & \textbf{8.68} & \textbf{6.23} & \textbf{16.12} & \textbf{16.56} & \textbf{14.85} \\
        \midrule
        \multirow{3}{*}{Horse} 
        & DDIM & 22.17 & 12.92 & 7.92 & 157.24 & 73.50 & 22.59 & 266.54 & 243.47 & 181.69 & 316.98 & 301.88 & 226.54 \\
         & DiffAE & 11.97 & 9.37 & 7.44 & 9.62 & 6.83 & 4.71 & 12.41 & 10.07 & 7.88 & 24.04 & 21.19 & 18.34 \\
         & \modelnamenew & 12.53 & 9.80 & 8.90 
         & \textbf{9.15} & \textbf{6.36} & \textbf{3.98} & \textbf{10.45}
         & \textbf{8.31} & \textbf{6.19} & \textbf{19.64} & \textbf{17.18} & \textbf{14.55} \\
         \midrule
         \multirow{5}{*}{Bedroom} & LDM & 41.67 & 10.86 & \textbf{4.39} & \grayline{\underline{2.28}} &  \grayline{5.80} & \graylineempty{67.42} \\
        & CM-CD & \grayline{7.01} & \grayline{29.53} & \grayline{169.57} & \graylineempty{241.90} \\
        & DDIM & 13.70 & 9.23 & 7.14 & 108.72 & 59.11 & 11.81 & 190.04 & 157.81 & 76.28 & 250.50 & 224.96 & 139.81 \\
         & DiffAE & \textbf{10.69} & 8.19 & 6.50 & 7.10 & 5.26 & 4.13 & 8.42 & 7.08 & 6.00 & 15.15 & 13.32 & 12.01 \\
         & \modelnamenew & 11.01 & \textbf{8.03} & 6.35 & \textbf{6.21} & \textbf{4.77} & \textbf{3.49} & \textbf{7.26} & \textbf{6.31} & \textbf{5.39} & \textbf{13.20} & \textbf{11.83} & \textbf{10.58} \\ 
    \bottomrule
    \end{tabular}}
    \caption{Overall performance comparison in various tasks. We use FID and rFID to evaluate the performances of the generation and reconstruction, respectively. For the interpolation tasks, we test two settings $\alpha=0.2$ and $\alpha=0.4$, where FID is also used to evaluate the performance. We highlight the best performance within the performances of methods that need to set up the parameter $T$ with {\bf bold}. The performances of the methods with only one step such as NVAE, StyleGAN-XL, and CM-CD are \uline{underlined} if their performances are better than the best performances of other methods with $T=50$.}
    \label{tab:overall_performance_comparison}
\end{table*}

\subsubsection{Image Generation}
\begin{figure}[t]
    \centering
    \begin{subfigure}{}
        \includegraphics[width=\textwidth]{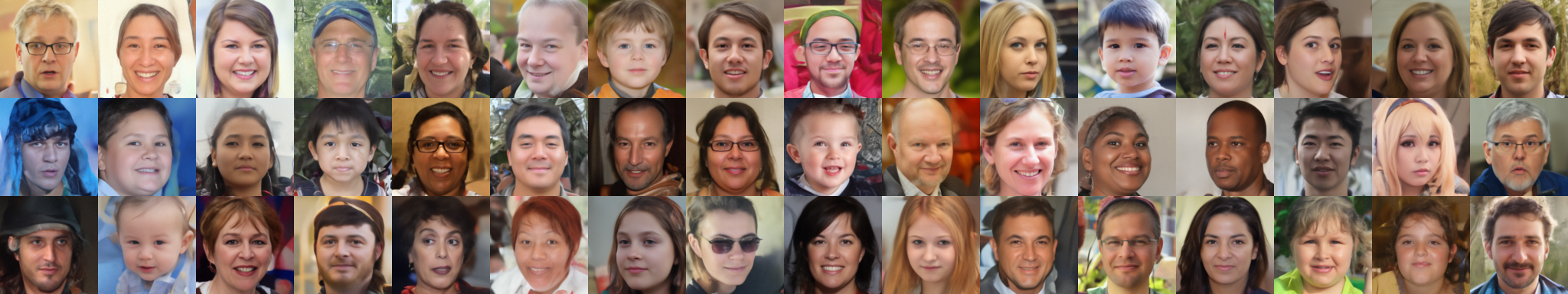}
        \caption{FFHQ128, T=10}
        \label{samples_t10}
        \includegraphics[width=\textwidth]{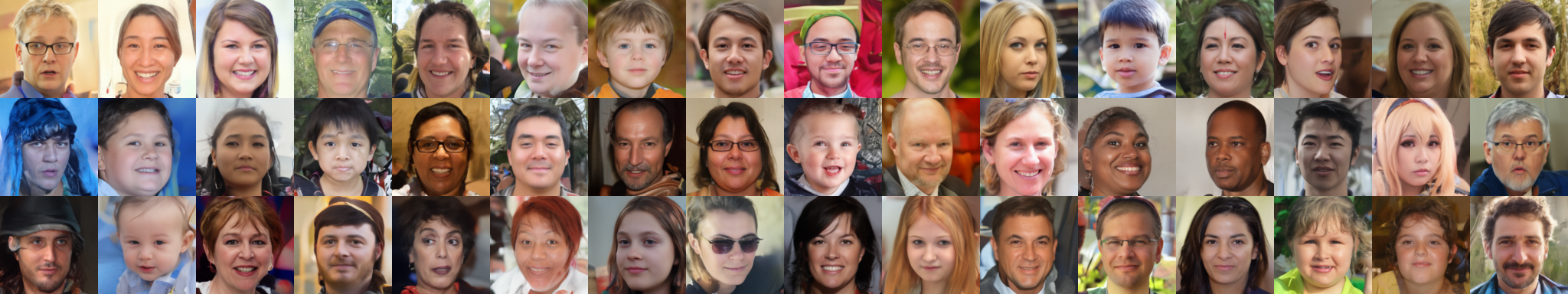}
        \caption{FFHQ128, T=20}
        \label{samples_t20}
        \includegraphics[width=\textwidth]{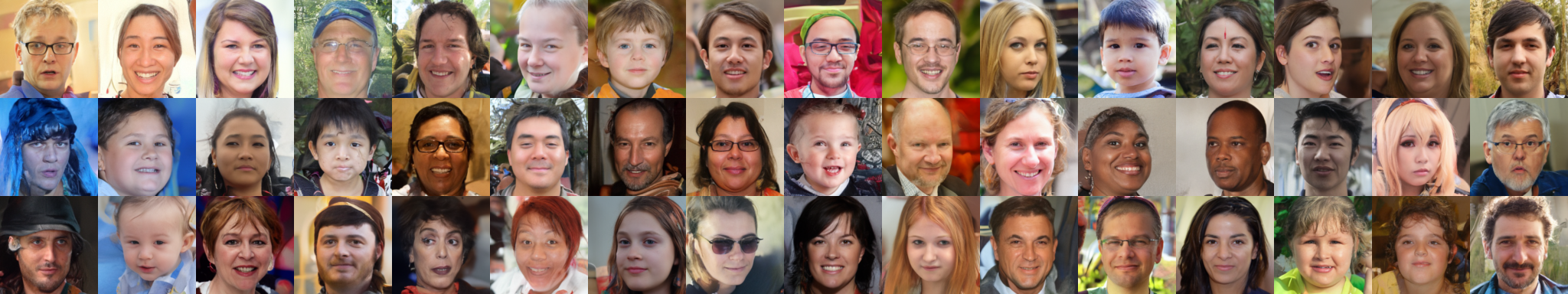}
        \caption{FFHQ128, T=50}
        \label{samples_t50}
    \end{subfigure}
\end{figure}

\begin{figure}[t]
    \centering
    \begin{subfigure}{} 
        \includegraphics[width=\textwidth]{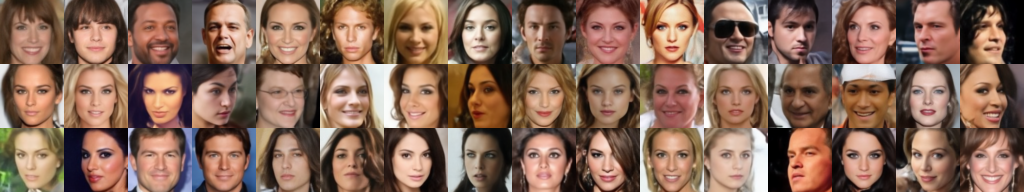}
        \caption{CelebA64, T=10}
        \label{celeba_samples_t10}
        \includegraphics[width=\textwidth]{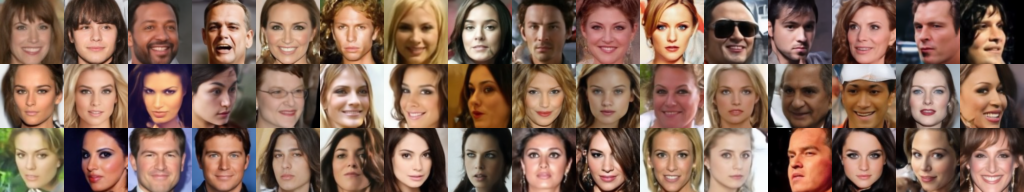}
        \caption{CelebA64, T=20}
        \label{celeba_samples_t20}
        \includegraphics[width=\textwidth]{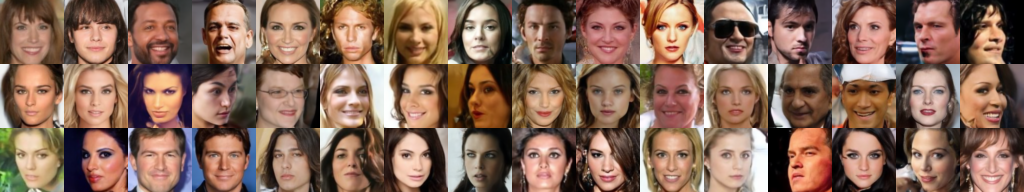}
        \caption{CelebA64, T=50}
        \label{celeba_samples_t50}
    \end{subfigure}
\end{figure}

For FFHQ128, we present the generated images in 
Figure \ref{samples_t10}, \ref{samples_t20}, \ref{samples_t50}). Then for CelebA64, the generated images are shown in Figure \ref{celeba_samples_t10}, \ref{celeba_samples_t20}, \ref{celeba_samples_t50}. As depicted in the figures, images generated with \( T = 50 \) typically exhibit finer granularity when contrasted against those produced with \( T = 10 \) or \( T = 20 \).

\subsubsection{Image Reconstruction}
\begin{figure}[t]
    \centering
    \includegraphics[width=\textwidth]{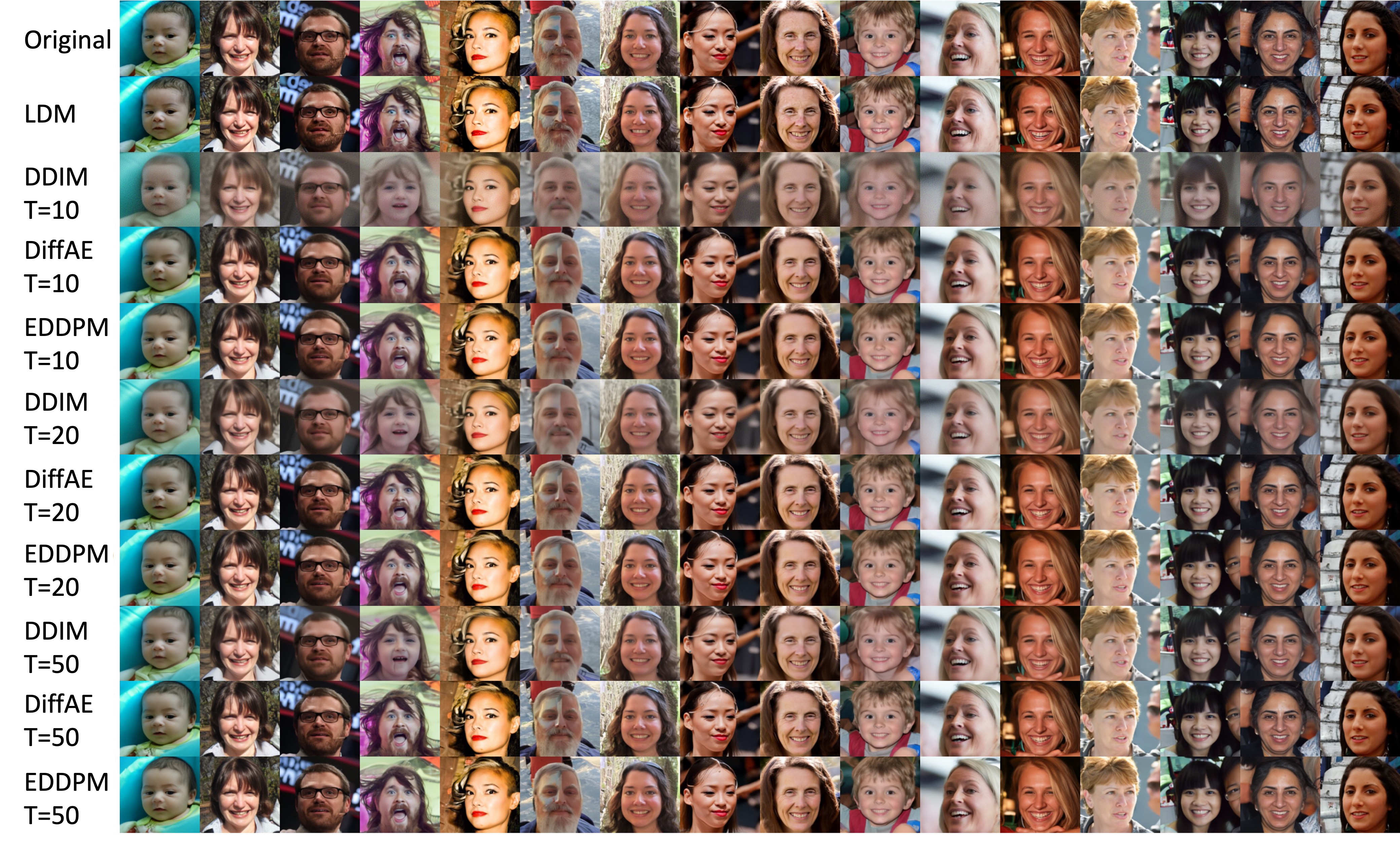}
    \caption{Image reconstructions for FFHQ128 with different models. Best viewed with zooming in.}
    \label{fig:ffhq_reconstruction}
\end{figure}

\begin{figure}[t]
    \centering
    \includegraphics[width=\textwidth]{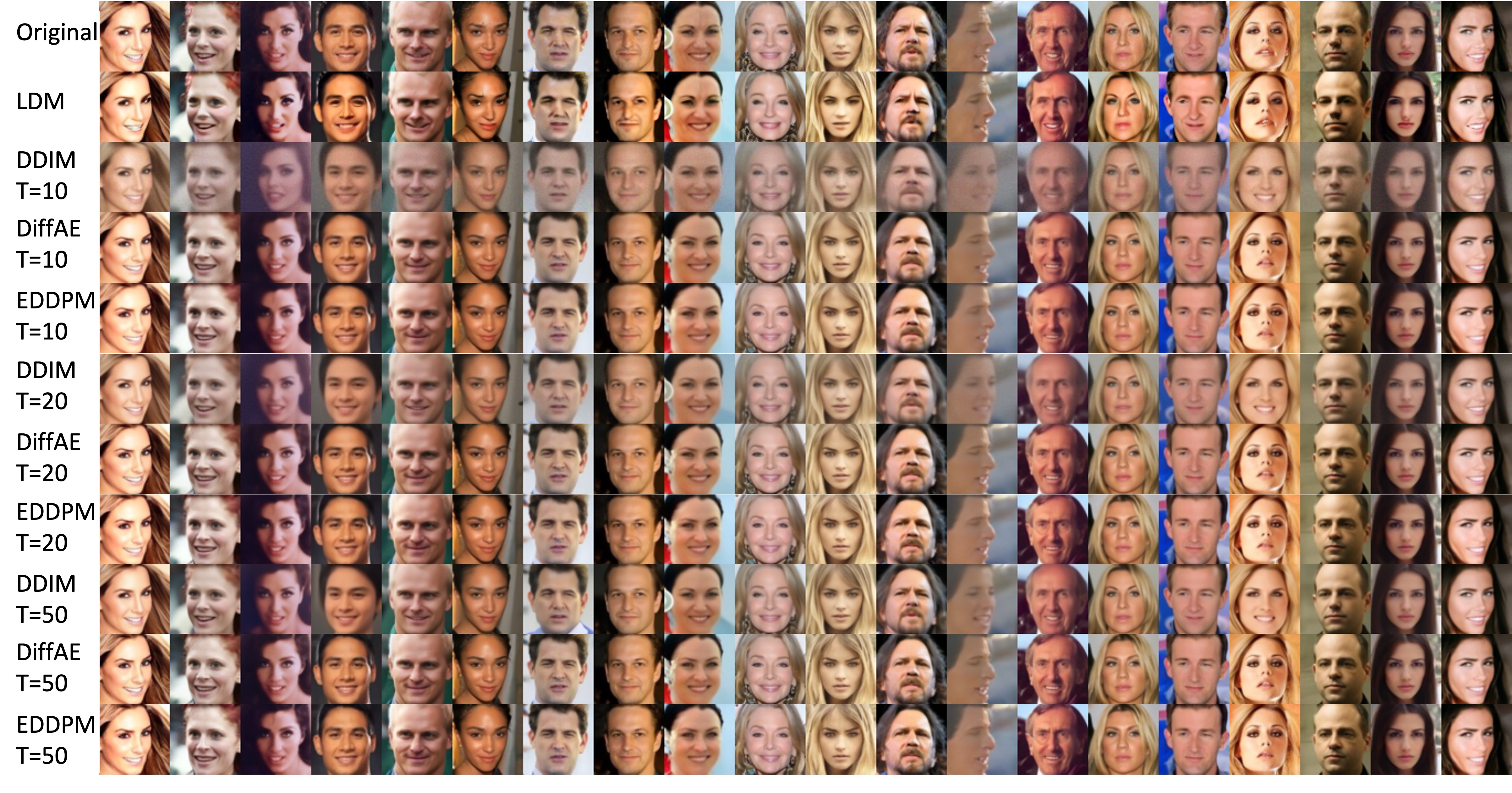}
    \caption{Image reconstructions for CelebA64 with different models. Best viewed with zooming in.}
    \label{fig:celeba_reconstruction}
    \vspace{-10pt}
\end{figure}

For a comparative analysis, we present reconstructed images from various models, each utilizing a distinct total step \( T \) in the decoder. The results for FFHQ128 and CelebA64 are depicted in Figure \ref{fig:ffhq_reconstruction} and Figure \ref{fig:celeba_reconstruction}, respectively.
From the figures, it's clear that the VAE in LDM is effective at reconstruction. Our model \modelnamenew also produces strong results.


\subsubsection{Image Representation}
\begin{figure}[t]
    \centering
    \includegraphics[width=\textwidth]{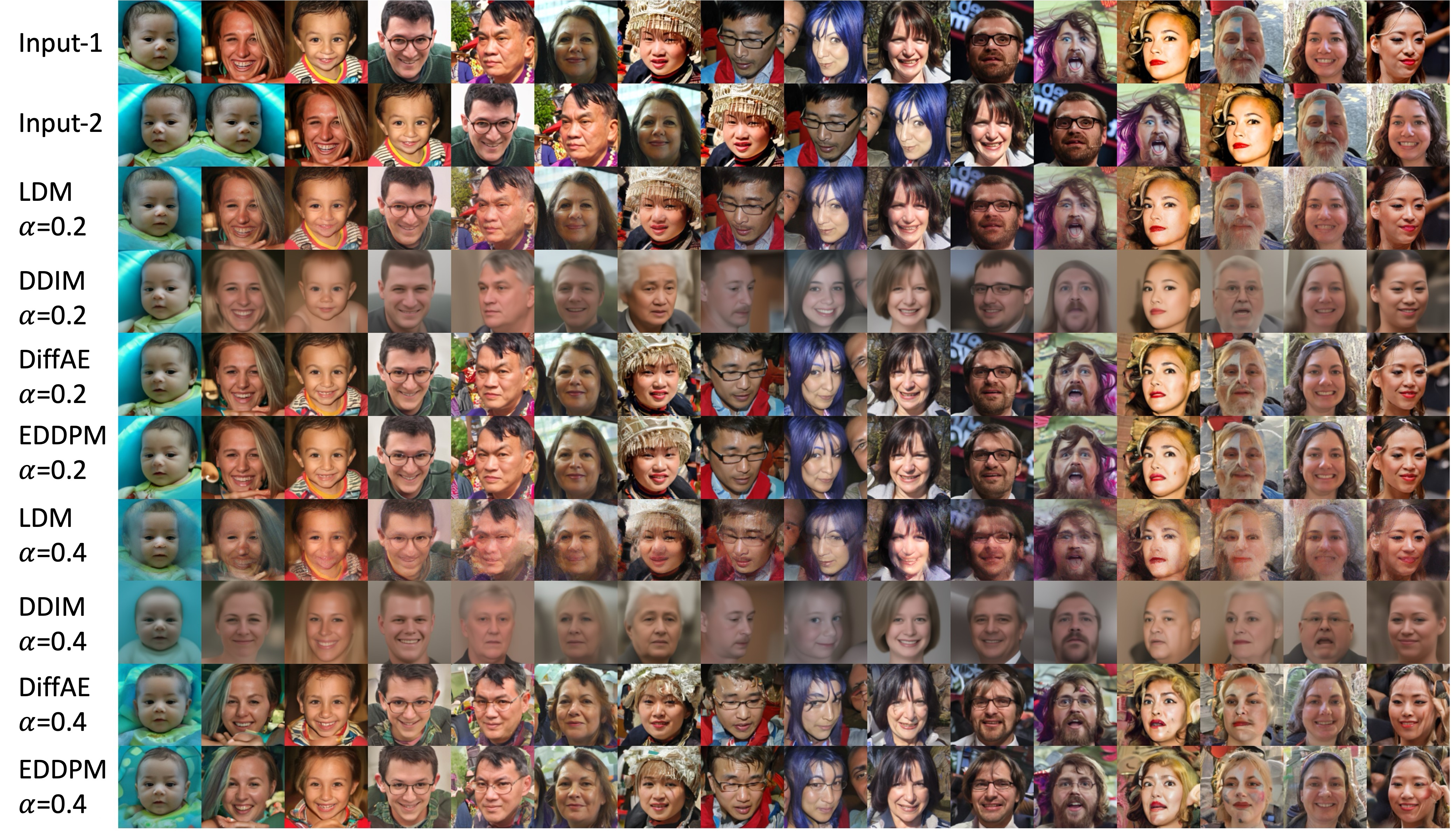}
    \caption{Image interpolations for FFHQ128 with different models. $T$ is fixed to 50 in these experiments. Best viewed with zooming
in.}
    \label{fig:ffhq128_representation}
\end{figure}

\begin{figure}[t]
    \centering
    \includegraphics[width=\textwidth]{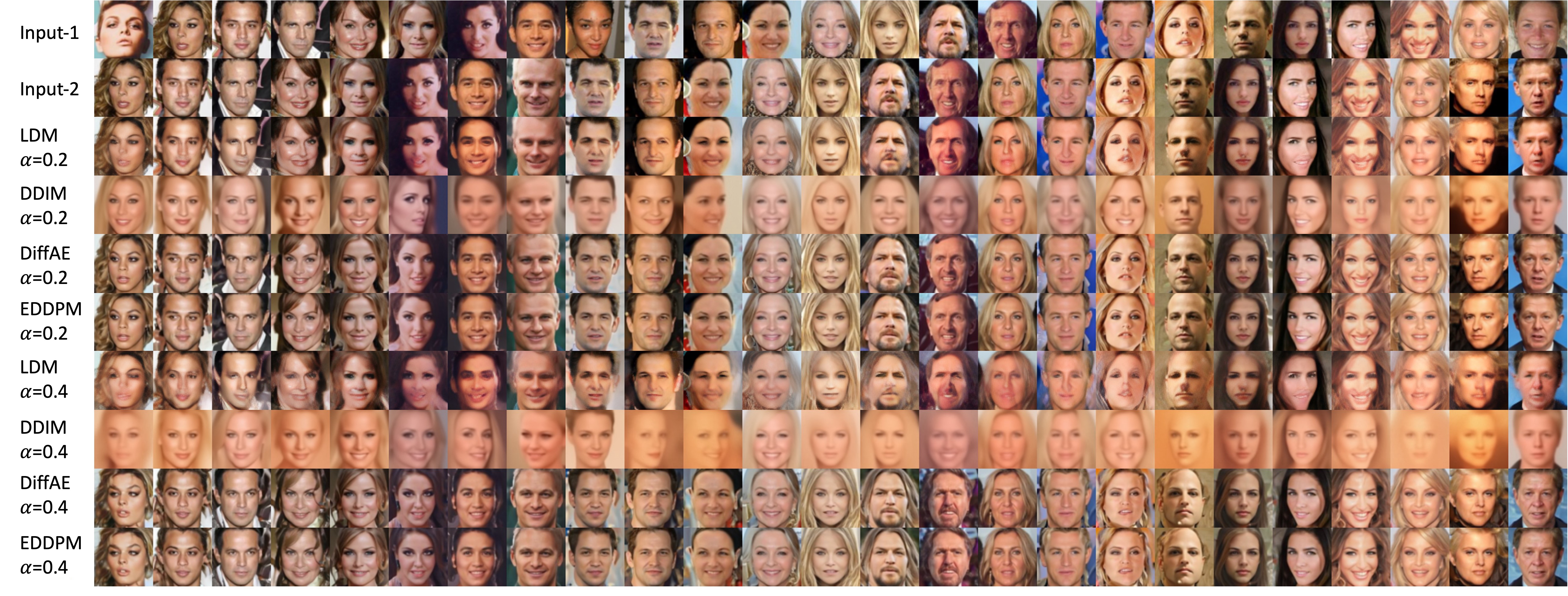}
    \caption{Image interpolations for CelebA64 with different models. $T$ is fixed to 50 in these experiments. Best viewed with zooming
in.}
    \label{fig:celeba_representation}
\end{figure}

\paragraph{Interpolation} 
The interpolation results for FFHQ128 and CelebA64 are illustrated in Figure \ref{fig:ffhq128_representation} and Figure \ref{fig:celeba_representation}, respectively. A close examination of the figures reveals that while the VAE in LDM is adept at reconstruction, its interpolation with \( \alpha = 0.4 \) appears akin to a superposition of two images. This aligns with the inherent nature of their VAE, where the representation predominantly encodes spatial rather than semantic information. In contrast, both our approach and DiffAE yield superior results at \( \alpha = 0.4 \). Specifically, our method demonstrates fewer visual artifacts compared with DiffAE, underscoring the better representation space of our model.

\paragraph{Manipulation} 
The comprehensive comparisons are presented in Table \ref{tab:manipulation_classification}. As indicated by the table, our model consistently delivers comparable AUC values across all classes.

\begin{table}[ht]
    \centering
    \small
    \begin{tabular}{cccc}
    \toprule
 \textbf{Class} & \# \textbf{Positives} & \textbf{DiffAE} & \textbf{\modelnamenew} \\
\midrule
5\_o\_Clock\_Shadow & 1314 & 0.9469 & 0.9466 \\
Arched\_Eyebrows & 3171 & 0.8822 & 0.8811 \\
Attractive & 5009 & 0.8849 & 0.8792 \\
Bags\_Under\_Eyes & 2528 & 0.8787 & 0.8821 \\
Bald & 174 & 0.9886 & 0.9843 \\
Bangs & 1593 & 0.9779 & 0.9770 \\
Big\_Lips & 3187 & 0.7031 & 0.7108 \\
Big\_Nose & 2796 & 0.8757 & 0.8683 \\
Black\_Hair & 1895 & 0.9483 & 0.9427 \\
Blond\_Hair & 1469 & 0.9775 & 0.9760 \\
Blurry & 27 & 0.8434 & 0.8762 \\
Brown\_Hair & 2074 & 0.8476 & 0.8334 \\
Bushy\_Eyebrows & 1658 & 0.9174 & 0.9117 \\
Chubby & 579 & 0.9439 & 0.9339 \\
Double\_Chin & 481 & 0.9503 & 0.9486 \\
Eyeglasses & 404 & 0.9916 & 0.9877 \\
Goatee & 650 & 0.9724 & 0.9690 \\
Gray\_Hair & 346 & 0.9865 & 0.9856 \\
Heavy\_Makeup & 3963 & 0.9714 & 0.9695 \\
High\_Cheekbones & 4043 & 0.9490 & 0.9474 \\
Male & 3184 & 0.9977 & 0.9969 \\
Mouth\_Slightly\_Open & 4125 & 0.9784 & 0.9777 \\
Mustache & 495 & 0.9573 & 0.9547 \\
Narrow\_Eyes & 1061 & 0.8569 & 0.8600 \\
No\_Beard & 7047 & 0.9784 & 0.9754 \\
Oval\_Face & 1772 & 0.7494 & 0.7469 \\
Pale\_Skin & 458 & 0.9635 & 0.9618 \\
Pointy\_Nose & 2738 & 0.7263 & 0.7235 \\
Receding\_Hairline & 718 & 0.9373 & 0.9340 \\
Rosy\_Cheeks & 975 & 0.9482 & 0.9416 \\
Sideburns & 684 & 0.9768 & 0.9726 \\
Smiling & 4050 & 0.9827 & 0.9803 \\
Straight\_Hair & 1866 & 0.8030 & 0.8027 \\
Wavy\_Hair & 3099 & 0.8804 & 0.8770 \\
Wearing\_Earrings & 2328 & 0.8933 & 0.8851 \\
Wearing\_Hat & 311 & 0.9875 & 0.9862 \\
Wearing\_Lipstick & 4911 & 0.9803 & 0.9789 \\
Wearing\_Necklace & 1496 & 0.7788 & 0.7789 \\
Wearing\_Necktie & 600 & 0.9583 & 0.9560 \\
Young & 6871 & 0.9230 & 0.9130 \\
\midrule
Weighted Avg AUC & - & \textbf{0.9174} & 0.9154 \\
Weighted Avg ACC & - & 0.7954 & \textbf{0.8929} \\
\bottomrule
    \end{tabular}
    \caption{Classification performance comparison.}
    \label{tab:manipulation_classification}
\end{table}

\subsection{Protein}
\label{app:exp_prot}
Protein design plays a crucial role in drug discovery, protein therapeutics and various other applications in biotechnology. However, due to the complex and large search space of protein sequences, traditional empirical methods that demands intensive and thorough experiments and screening for validation are expensive and time-consuming. Recent advances in machine learning and computational methods introduces new approaches for protein optimization, however, most of these methods focuses on protein design in the discrete text space and lacks a meaningful latent space. Our work combines the Autoencoder and Diffusion Model, enabling effective protein generation/optimization and establishing a robust latent space for protein representation. 

\subsubsection{Setup}
\label{app:exp_prot:setup}
\paragraph{Architecture}
We adopt the set up of ReLSO \citep{castro2022relso} which consists of a simple transformer as the encoder and convolutional layers as the decoder. While ReLSO relies on Negative Sampling Loss which augments the datasets with synthetic samples with negative labels and Interpolative Regularization which penalizes differences between interpolated points and nearest neighbors to achieve a smooth pseudo-convex latent space, we leverage the same MLP with skip connections in our text model as the backbone of the diffusion model to act as a learnable prior. In line with ReLSO, we jointly train a simple regressor consists of linear layers and ReLU that predicts the fitness value with the latent embedding of a protein sequence. The regressor is trained with Mean Squared Error and added to final objective derived in \S\ref{app:deriv:our_final} : 
\begin{align}
    \mathcal L_t^{\text{protein}} = {w_{\text{protein}} * \mathcal L^{\text{final}}}
     + \mathcal L^{\text{regressor}}, \label{eq:protein_loss} 
\end{align}
The dimension of the latent space is set to 30. And we have trained 2 separate models with $w_{\text{protein}}$ set to 1 and 5.

\paragraph{Dataset}
The models are trained and evaluaed on the Gifford \citet{10.1093/bioinformatics/btz895} dataset and the GFP\citep{gfp} dataset. The Gifford dataset was generated from directed evolution of $10^{10}$ mutants of an antibody against a single target(Ranibizumab) through three rounds of phage display panning. Fitness value is defined as the log of the round-to-round ratio of sequence frequencies and a higher fitness value indicate better performance. In this dataset, each protein sequence has a length of 20 with a vocabulary of 20 amino acids which are represented by 20 letters. The resulting dataset consists of 57603 sequences in the training set, 10166 sequences in the validation set, and 22690 sequences in the test set. The GFP dataset was sampled from a fluorescent protein(avGFP) through mutagenesis. The dataset consists of 51175 sequences with length of 237 and an average mutation of 3.7. The fitness is defined as the fluorescence of the proteins assayed with fluorescence-activated cell sorting of the sequences. 

\paragraph{Baseline}
In addition to ReLSO, an autoencoder-based model introduced above, we trained a vanilla Variational Autoencoder\citep{kingma2013auto} consists of the same transformer as encoder and convolutional layers as decoder. The dimension of the latent space of ReLSO and VAE are all set to 30. Furthermore, we have conducted comparisons with latest advancements in protein sequence diffusion models, namely NOS \citep{gruver2023protein}. NOS proposes a Discrete Diffusion Model that based on BERT and utilized [MASK] as noise, and a Gaussian Diffusion Model that jointly learned a word embedding function and used BERT as the backbone of the diffusion model. While ReLSO, VAE and \modelnamenew utilized a regressor of linear layers and ReLU, the regressor of NOS required an additional BertPooler to aggregate a sequence of latent embeddings. We have conducted experiments on Protein Representation with \modelnamenew, ReLSO, VAE and NOS; Protein Optimization with \modelnamenew, ReLSO and VAE; Protein Reconstruction with \modelnamenew, ReLSO and VAE; and Protein Generation with \modelnamenew, VAE and NOS. 

\begin{table}
    \centering
    \small
    \vspace{-5pt}
    \begin{tabular}{c|c|ccccc}
        Dataset & Model & MSE & L1 &  Pearson & Spearman & Reconstruction CE\\
    \midrule
         \multirow{6}{*}{Gifford} 
         & VAE~\cite{kingma2013auto}             & 0.255 & 0.361 & 0.829 & 0.454 & 1.464\\
         & ReLSO~\cite{castro2022relso}          & 0.293 & 0.401 & 0.826 & 0.477  & 0.940\\
         & NOS-Discrete~\cite{gruver2023protein} & 0.563 & 0.648 & 0.633 & 0.407  & - \\
         & NOS-Gaussian~\cite{gruver2023protein} & 0.753 & 0.785 & 0.350 & 0.292 & - \\
         & \modelnamenew($w_{\text{protein}}$=0.5)                     & 0.231 & 0.346 & 0.840 & 0.477  & 1.016\\
         & \modelnamenew($w_{\text{protein}}$=5)                       & 0.211 & 0.331 & 0.844 & 0.472  & 1.143\\
    \midrule
        \multirow{6}{*}{GFP} 
         & VAE ~\cite{kingma2013auto}            & 0.681 & 0.615 & 0.873  & 0.764  & 0.093\\
         & ReLSO~\cite{castro2022relso}          & 0.516 & 0.592 & 0.793  & 0.701 & 0.099\\
         & NOS-Discrete~\cite{gruver2023protein} & 7.984 & 2.620 & 0.378  & 0.350 & -\\
         & NOS-Gaussian~\cite{gruver2023protein} & 8.040 & 2.630 & -0.124 & -0.129 & -\\
         & \modelnamenew($w_{\text{protein}}$=0.5)                     & 0.496 & 0.464 & 0.854  & 0.753  & 0.095\\
         & \modelnamenew($w_{\text{protein}}$=0.5)                       & 0.488 & 0.440 & 0.848  & 0.734  & 0.094\\
    \bottomrule
    \end{tabular}
    \caption{Protein Representation and Reconstruction}
    \vspace{-5pt}
    \label{tab:protein_representation}
\end{table}

\subsubsection{Protein Optimization}
\label{app:exp_prot:opt}
We optimize a protein sequence by optimizing its corresponding embedding in the latent space. Given a protein sequence $\rvx_0$, we first obtain its latent embedding with $ \rvx_1 = \mathcal E_\lambda(\rvx_0) $. We adopt the sampling algorithm introduced in LatentOps that solves an ODE involving the regressor \citet{liu2022composable}. This approach requires a target fitness value to guide the optimization. We set this value to 1, 1.5, 2, and 2.5 for the Gifford dataset and 3, 4 for the GFP dataset which are all reasonably high fitness values in their corresponding datasets. For evaluation, we optimize 60 random protein sequence and evaluate their fitness value using the regressor. We assess the results on Diversity, quantified by the average Levenshtein distance of each sequence relative to the other 59 optimized sequences; on Novelty, determined by the median of the minimum Levenshtein distance between each optimized sequence and the training set; and on Quality, measured by the negative log likelihood given by ProtGPT2, a large protein language model trained on the UniRef50 dataset. The results are presented in Table \ref{tab:protein_optimization}, Table \ref{tab:more_protein_optimization_gifford} and Table \ref{tab:more_protein_optimization_gfp}.

\begin{table}[t]
    \centering
    \small
    \begin{tabular}{c|cccccc}
        Dataset & Model  & Max Fitness & Mean Fitness & Diversity & Novelty & NLL \\
    \midrule
        \multirow{4}{*}{Gifford} 
        &VAE~\cite{kingma2013auto}    & 2.005 & 1.969  & 14.789 & 7 & 29.301 \\
        &ReLSO~\cite{castro2022relso} & 0.738 & 0.737  & 2.14  & 7 & 28.736 \\
        &\modelnamenew ($w_{\text{protein}}$=0.5)         & 2.003 &	2.000  & 12.615 & 6 & 29.219 \\
        &\modelnamenew ($w_{\text{protein}}$=5)           & 2.000 &	2.000  & 12.256 & 4 & 29.452 \\
    \midrule
        \multirow{4}{*}{GFP}
        &VAE~\cite{kingma2013auto}     & 4.007 & 4.000 & 2.743  & 1 & 23.476 \\
        &ReLSO~\cite{castro2022relso}  & 2.875 & 2.524 & 1.70     & 1 & 23.468 \\
        &\modelnamenew ($w_{\text{protein}}$=0.5)          & 4.001 & 4.000 & 12.836 & 1 & 23.488 \\
        &\modelnamenew ($w_{\text{protein}}$=5)            & 4.001 & 4.000 & 1.076  & 1 & 23.480 \\
    \bottomrule
    \end{tabular}
    \caption{Comparison of Protein Optimization.}
    \label{tab:protein_optimization}
\end{table}

\begin{table}[t]
    \centering
    \small
    \begin{tabular}{c|cccccc}
        Target Fitness & Model  & Max Fitness & Mean Fitness & Diversity & Novelty & NLL \\
    \midrule
        \multirow{2}{*}{1} 
        &ReLSO                                    & 0.738 & 0.737  & 2.70 & 6 & 28.645 \\
        &\modelnamenew ($w_{\text{protein}}$=0.5) & 1.000 & 1.000  & 13.35  & 6 & 29.352 \\
    \midrule
        \multirow{2}{*}{1.5} 
        &ReLSO                                    & 0.738 & 0.737  & 2.06 & 7 & 28.713 \\
        &\modelnamenew ($w_{\text{protein}}$=0.5) & 1.501 & 1.500  & 13.07  & 6 & 29.294 \\
    \midrule
        \multirow{2}{*}{2} 
        &ReLSO                                    & 0.738 & 0.737  & 2.14 & 7 & 28.736 \\
        &\modelnamenew ($w_{\text{protein}}$=0.5) & 2.003 & 2.000  & 12.61  & 6 & 29.219 \\
    \midrule
        \multirow{2}{*}{2.5} 
        &ReLSO                                    & 0.738 & 0.737  & 2.26 & 7 & 28.790 \\
        &\modelnamenew ($w_{\text{protein}}$=0.5) & 2.504 & 2.500  & 12.30  & 6 & 28.736 \\
    \bottomrule
    \end{tabular}
    \caption{Comparison of Protein Optimization with different Target Fitness Value on Gifford.}
    \label{tab:more_protein_optimization_gifford}
\end{table}

\begin{table}[t]
    \centering
    \small
    \begin{tabular}{c|cccccc}
        Target Fitness & Model  & Max Fitness & Mean Fitness & Diversity & Novelty & NLL \\
    \midrule
        \multirow{2}{*}{3} 
        &ReLSO                                    & 2.873 & 2.872  & 1.70 & 1 & 23.468 \\
        &\modelnamenew ($w_{\text{protein}}$=0.5) & 3.001 & 3.000  & 3.41  & 1 & 23.478 \\
    \midrule
        \multirow{2}{*}{4} 
        &ReLSO                                    & 2.873 & 2.872  & 1.08 & 1 & 23.469 \\
        &\modelnamenew ($w_{\text{protein}}$=0.5) & 4.000 & 4.000  & 12.83  & 1 & 23.488 \\
    \bottomrule
    \end{tabular}
    \caption{Comparison of Protein Optimization with different Target Fitness Value on GFP.}
    \label{tab:more_protein_optimization_gfp}
\end{table}

\begin{table}[t]
    \centering
    \small
    \begin{tabular}{c|cccccc}
        Dataset & Model  & Max Fitness & Mean Fitness & Diversity & Novelty & NLL \\
    \midrule
        \multirow{5}{*}{Gifford} 
        &VAE~\cite{kingma2013auto}               & 1.010 & 0.999  & 15.771 & 9 & 29.389 \\
        & NOS-Discrete~\cite{gruver2023protein}  & 1.650 & 0.702  & 14.549 & 10 & 29.01 \\
        & NOS-Gaussian~\cite{gruver2023protein}  & 0.493 & -0.112  & 8.160 & 3 & 29.05 \\
        &\modelnamenew ($w_{\text{protein}}$=0.5)                    & 1.003 &	1.000  & 13.067 & 6 & 29.200 \\
        &\modelnamenew ($w_{\text{protein}}$=5)                      & 1.004 &	1.000  & 12.486 & 5 & 29.490 \\
    \midrule
        \multirow{5}{*}{GFP}
        &VAE~\cite{kingma2013auto}               & 3.036 & 3.000  & 114.2  & 105.5 & 23.702 \\
        & NOS-Discrete~\cite{gruver2023protein}  & 0.056 & 0.023  & 22.450 & 220  & 10.218 \\
        & NOS-Gaussian~\cite{gruver2023protein}  & -0.005 & -0.007 & 3.626 & 220 & 10.066 \\
        &\modelnamenew ($w_{\text{protein}}$=0.5)                    & 3.007 & 3.001  & 4.173  & 1 & 23.481    \\
        &\modelnamenew ($w_{\text{protein}}$=5)                      & 3.003 & 3.002  & 0.715  & 1 & 23.475    \\
    \bottomrule
    \end{tabular}
    \caption{Comparison of Protein Conditional Generation.}
    \label{tab:cond_protein_generation}
\end{table}

\subsubsection{Protein Reconstruction}
\label{app:exp_prot:rec}
Reconstruction is evaluated with Cross Entropy to compare input and reconstructed protein sequences in the test set. The results are presented in the last column of Table \ref{tab:protein_representation}. From Table \ref{tab:protein_representation}, \modelnamenew achieves comparable reconstruction ability compared to the baselines. Note that since NOS is a transformer-based model with word embedding that could not perform reconstruction in the same settings as other models. 

\subsubsection{Conditional Protein Generation} \label{app:exp_prot:gen}
For conditional generation with VAE and \modelnamenew, latent representations are sampled and optimized with LatentOps algorithm with target fitness of 1 and 3 for Gifford and GFP dataset respectively. For NOS, the latents are updated through out each step of the diffusion process with Langevin Dynamics using gradients of the regressor. The quantitative evaluation are presentsed in \ref{tab:cond_protein_generation}. Although NOS models achieved better NLL, the generated proteins are have lower fitness values than other methods. 

\subsubsection{Influence of weight $w$}
In the objective (Eq.\ref{eq:detail_final_loss}), the weight $w$ aims to balance the generation and reconstruction. As mentioned in \S\ref{sec:weight_w}, we adopt $w=8$ for text and $w=1$ for both image and protein. The choice of $w$ is primarily influenced by the different loss functions used for the reconstruction term $L_{Rec}$ (Eq.\ref{eq:ddpm_loss_kl}) across various data types. For example, cross-entropy is used for text, while mean squared error (MSE) is used for images. To ensure that the different loss terms have similar scales, we selected $w$ values that roughly balanced their magnitudes. The $w$ values are then fixed during training. As a comparison, beta-VAE on text typically requires careful scheduling/annealing of $\beta$ values during training~\cite{DBLP:conf/iclr/HigginsMPBGBML17,DBLP:conf/emnlp/LiGLPLZG20}, making the training more difficult and unstable.
\subsubsection{Posterior Collapse}
\modelnamenew can alleviate the \textit{posterior collapse} issue on text sequences thanks to the improved formulation and training:
\begin{itemize}
    \item As discussed in \S\ref{sec:method:learnable-prior}, \modelnamenews can be viewed as a VAE with a jointly learned diffusion model prior (instead of a standard Gaussian prior in vanilla VAEs). More formally, the \modelnamenews' objective in Eq.\ref{eq:our_loss_kl} includes the $L_\text{align}$ term that explicitly encourages alignment between the encoder's posterior distribution $q_\lambda(\mathbf x_1|\mathbf x_0)$ and the diffusion model's prior $p_\theta(\mathbf x_1|\mathbf x_2)$. This learned prior is more flexible and helps avoid the unreasonable regularization posed by standard Gaussian prior that causes posterior collapse.
    \item \modelnamenews derive the training objective from the well-established DDPM framework, enabling stable and effective joint training of the encoder, decoder, and diffusion components. For example, as mentioned above, \modelnamenews are robust to the balancing weight $w$. This is in contrast to the instability and sensitivity to hyperparameters in text-VAE training. To alleviate posterior collapse, training text-VAEs often requires tricks like beta-annealing~\cite{DBLP:conf/conll/BowmanVVDJB16}, free bits~\cite{kingma2017improving}, cyclic annealing schedule~\cite{fu2019cyclical}, and so on. This can be observed through the training loss curves in Figure \ref{fig:training_curves}.
\end{itemize}
Therefore, the \textit{posterior collapse} issue is not obvious in \modelnamenews.

\clearpage

\section{Contrasting with Similar Works}
In \S\ref{sec:method:connections}, we provide a high-level comparison of \modelnamenews with many related models, highlighting the connections and differences between our approach and various existing methods. In \S\ref{sec:related_work}, we present additional discussion of related works, focusing on recent efforts to combine VAEs, GANs, and diffusion models. We further mention the limitations of these approaches and how \modelnamenews aim to overcome them. In this section, we would like to offer an in-depth discussion of \modelnamenews with two closely related methods: Latent Diffusion Models and Latent Score-based Generative Models.
\subsection{Latent Diffusion Models}
\label{app:LDM}
Latent Diffusion Models (LDMs), often termed Stable Diffusion \cite{Rombach2021HighResolutionIS}, utilize an architectural combination of an autoencoder and a diffusion model in latent space. In this section, we delineate the primary distinctions between LDMs and our proposed approach in detail:
\paragraph{Autoencoder's Functionality:}
\textit{LDMs:} The overarching objective of their autoencoder is twofold: to compress images into a compact latent representation and to ensure robust reconstruction capabilities from these latent vectors. To bias the autoencoder towards stronger reconstruction, they introduce a KL term but assign it a minute KL weight ($\sim10^{-6}$). Their employment of a purely convolution-based encoder-decoder emphasizes spatial preservation in the latent space, which, while bolstering reconstruction, poses challenges to distilling semantically rich latent features.

\textit{Our Approach:} Our autoencoder extends beyond mere dimensionality reduction. It's intricately tailored to synchronize effectively with the diffusion process, thereby fostering a more semantically-coherent latent space. Instead of relying on the conventional KL regularization against a standard Gaussian, we harness a learnable prior, rendering our latent space more adaptive and insightful.

\paragraph{Training Paradigm:}
\textit{LDMs:} Their training strategy bifurcates into two discrete phases: initial autoencoder training followed by subsequent training of the diffusion model in latent space. Consequently, the latent space's architecture predominantly adheres to the objectives set forth by the autoencoder.

\textit{Our Approach:} Our methodology pivots on end-to-end training, ensuring the latent space's structure is sculpted by the holistic objectives of the entire model. This integrated approach instills the latent space with nuanced semantics and more discernible significance, enhancing both interpretability and utility.

\subsection{Latent Score-based Generative Model}
The Latent Score-based Generative Model (LSGM) enhances generative capabilities by learning Score-based Generative Models (SGM) within the latent space of a Variational Autoencoder (VAE). We will delineate the detailed differences between LSGM and \modelnamenews in the following sections.
\paragraph{Training Objective:}
\textit{LSGM:} Their method can be regarded as a Variational Autoencoder (VAE) with a Score-based Generative Model (SGM) as the prior. Consequently, their objective is derived from \eqref{eq:our_vae_deriv}. However, they uniquely decompose the KL term into two components, and each component is approximated based on certain assumptions. This approach results in a fundamentally different final training objective to that of \modelnamenews.

\textit{Our Approach:} Our method begins with a foundational Diffusion model framework. In this context, we interpret the encoder and decoder as an extended step in the diffusion process. We then formulate a novel, generalized diffusion process that incorporates a learnable noise addition step. The derivation of our training objective strictly adheres to the conventional principles of standard diffusion models and does not rely on further approximations.

\paragraph{Training Paradigm:}
\textit{LSGM:} In their approach, there are two distinct training objectives: one for the VAE and another for the SGM. Although both losses are computed and used to update the model parameters, they are theoretically independent. For instance, the SGM objective does not update the weights of the VAE. This implies that the latent space of the VAE is not actively regularized by the SGM; rather, the SGM learns the latent distribution as defined by the VAE. Furthermore, the training process for their model incorporates some complex tricks, resulting in instability. This complexity makes it challenging to train their model effectively, a point acknowledged by the authors themselves in their repository \url{https://github.com/NVlabs/LSGM?tab=readme-ov-file#common-issues}.

\textit{Our Approach:} Contrarily, our method employs a singular, unified training objective for the entire model, which is trained in an end-to-end manner. The latent diffusion model is learned in conjunction with the encoder/decoder, which actively regularizes the latent space. This results in an improved latent structure, as detailed in \S\ref{sec:method}. 

\end{document}